\documentclass[runningheads]{llncs}

\usepackage{eccv}

\usepackage{amsmath,amsfonts,bm}

\def\eqref#1{equation~\ref{#1}}

\def\1{\bm{1}}

\DeclareMathAlphabet{\mathsfit}{\encodingdefault}{\sfdefault}{m}{sl}
\SetMathAlphabet{\mathsfit}{bold}{\encodingdefault}{\sfdefault}{bx}{n}

\usepackage{amsmath}
\usepackage{amsfonts}
\usepackage{amssymb}
\usepackage{wrapfig}
\usepackage{subcaption}
\usepackage{multirow}
 \usepackage{mathtools} 

\usepackage{verbatim}

\usepackage{anyfontsize}

\usepackage{microtype}
\usepackage{graphicx}
\usepackage{booktabs} 
\usepackage{multirow}
\usepackage{amsmath,amssymb}
\usepackage{booktabs}
\usepackage{caption,subcaption}

\usepackage{xcolor}
\definecolor{mygreen}{HTML}{3cb44b}
\definecolor{skyblue}{HTML}{beffff}
\definecolor{lightgreen}{HTML}{90ee90}

\usepackage{color, colortbl}

\definecolor{emerald}{rgb}{0.31, 0.78, 0.37}

\usepackage{tcolorbox}
\usepackage{enumitem}
\setitemize{itemsep=10pt,topsep=0pt,parsep=0pt,partopsep=0pt}
\usepackage{colortbl}

\usepackage{xcolor}
\definecolor{mygreen}{HTML}{3cb44b}
\colorlet{myyellow}{green!10!orange!90!}
\makeatletter

\usepackage{tikz}
\usetikzlibrary{arrows,shapes,snakes,automata,backgrounds,fit,petri}
\usepackage{adjustbox}

\newcommand{\RN}[1]{%
	\textup{\lowercase\expandafter{\it \romannumeral#1}}%
}
\usepackage{tabu}

\newcommand{\beq}{\vspace{0mm}\begin{equation}}
\newcommand{\eeq}{\vspace{0mm}\end{equation}}
\newcommand{\beqs}{\vspace{0mm}\begin{eqnarray}}
\newcommand{\eeqs}{\vspace{0mm}\end{eqnarray}}
\newcommand{\barr}{\begin{array}}
\newcommand{\earr}{\end{array}}

\newcommand{\Imat}{{\bf I}}

\newcommand{\Xmat}[0]{{{\bf X}}}

\usepackage{color, colortbl}
\definecolor{Gray}{gray}{0.93}

\usepackage{lipsum}

\usepackage{pifont}

\usepackage{makecell}

\usepackage{xcolor,amsmath}
\usepackage[linesnumbered,ruled,vlined]{algorithm2e}
\DontPrintSemicolon

\usepackage{xcolor}
\definecolor{mygreen}{HTML}{3cb44b}

\SetKwComment{Comment}{\color{green!50!black}\# }{}

\SetKwProg{Function}{def}{:}{}

\SetKwProg{For}{for}{:}{}
\SetKwProg{If}{if}{:}{}

\usepackage{eccvabbrv}

\usepackage{graphicx}
\usepackage{booktabs}

\usepackage[accsupp]{axessibility}  

\usepackage{hyperref}

\usepackage{orcidlink}
\usepackage[utf8]{inputenc}

\usepackage{array, caption, floatrow, makecell, booktabs}
\usepackage{wrapfig}
\usepackage{floatrow}
\usepackage{comment}
\usepackage{float}
\usepackage{wrapfig,lipsum,booktabs}
\usepackage{tabularx} 
\usepackage{tabu}

\usepackage[capitalize]{cleveref}
\crefname{section}{Sec.}{Secs.}
\Crefname{section}{Section}{Sections}
\Crefname{table}{Table}{Tables}
\crefname{table}{Tab.}{Tabs.}

\usepackage{multirow,array}
\usepackage{xcolor}
\usepackage{color, colortbl}
\definecolor{Gray}{gray}{0.93}

\newlength\savewidth\newcommand\shline{\noalign{\global\savewidth\arrayrulewidth
		\global\arrayrulewidth 1pt}\hline\noalign{\global\arrayrulewidth\savewidth}}

\newcommand\extrafootertext[1]{%
    \bgroup
    \renewcommand\thefootnote{\fnsymbol{footnote}}%
    \renewcommand\thempfootnote{\fnsymbol{mpfootnote}}%
    \footnotetext[0]{#1}%
    \egroup
}

\usepackage{listings}
\usepackage{xcolor}

\begin{document}

\title{Grounding DINO: Marrying DINO with Grounded Pre-Training for Open-Set Object Detection}

\titlerunning{Grounding DINO}

\author{Shilong Liu\textsuperscript{\rm 1,2}\thanks{This work was done when Shilong Liu, Feng Li, Hao Zhang, Jie Yang, and Qing Jiang were interns at IDEA.},\; Zhaoyang Zeng\textsuperscript{\rm 2}, Tianhe Ren\textsuperscript{\rm 2}, Feng Li\textsuperscript{\rm 2, 3}, Hao Zhang\textsuperscript{\rm 2, 3}, \\
Jie Yang\textsuperscript{\rm 2, 4}, Qing Jiang\textsuperscript{\rm 2, 6}
Chunyuan Li\textsuperscript{\rm 5}, Jianwei Yang\textsuperscript{\rm 5}, \\
Hang Su\textsuperscript{\rm 1}, Jun Zhu\textsuperscript{\rm 1}$^{\star\star}$, Lei Zhang\textsuperscript{\rm 2}\thanks{Corresponding authors.}.}

\authorrunning{S. Liu et al.}

\institute{\textsuperscript{\rm 1} Dept. of Comp. Sci. and Tech., BNRist Center, State Key Lab for Intell. Tech. \& Sys., \\
    Institute for AI, Tsinghua-Bosch Joint Center for ML, Tsinghua University \\
    \textsuperscript{\rm 2} International Digital Economy Academy (IDEA)\\
    \textsuperscript{\rm 3} The Hong Kong University of Science and Technology \\
    \textsuperscript{\rm 4} The Chinese University of Hong Kong (Shenzhen) \quad
    \textsuperscript{\rm 5} Microsoft Research, Redmond\\
    \textsuperscript{\rm 6} South China University of Technology\\
    \texttt{liusl20@mails.tsinghua.edu.cn, leizhang@idea.edu.cn}}

\maketitle

\begin{center}
    \centering
    \includegraphics[width=0.8\textwidth]{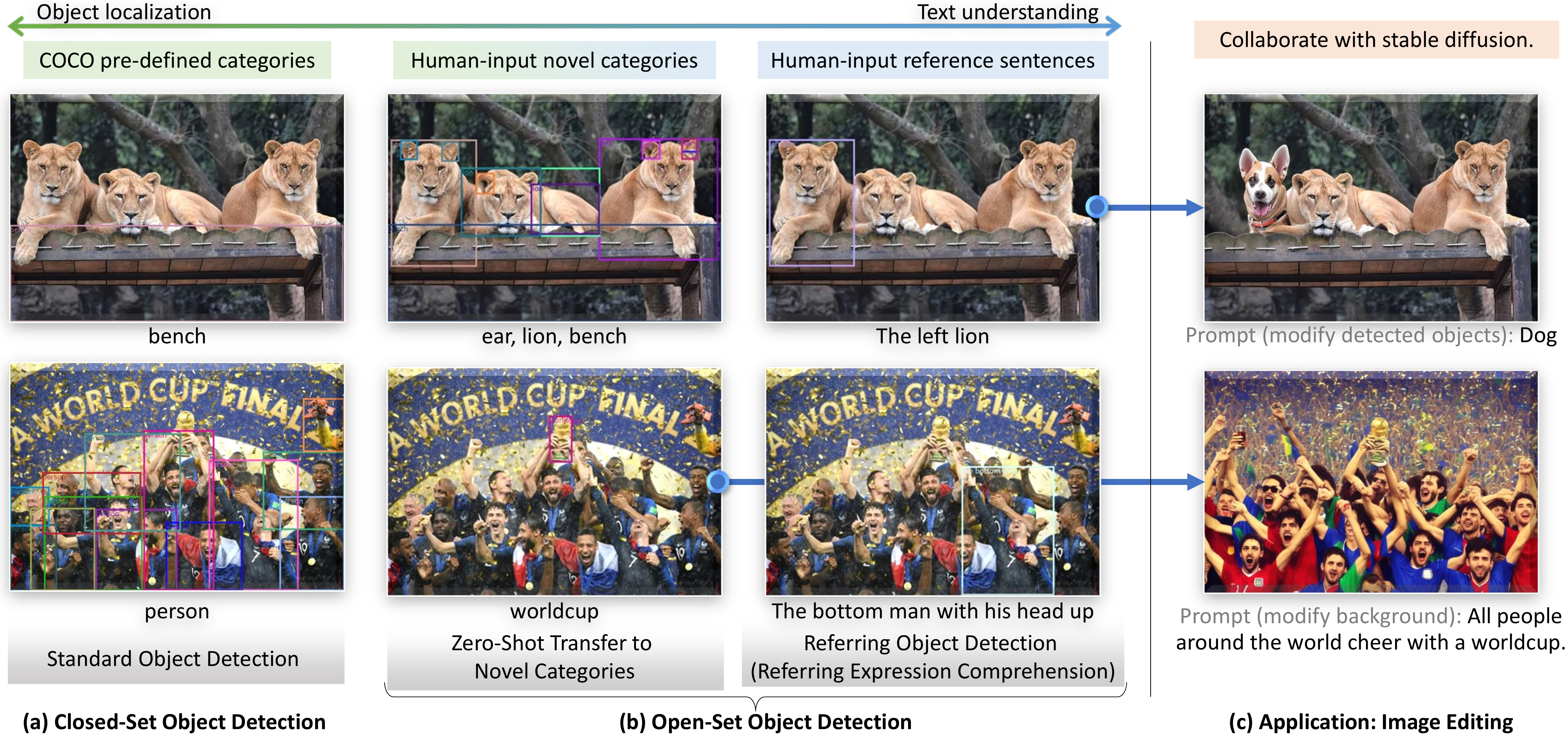}
    \vspace{-0.2cm}
    \captionof{figure}{(a) Closed-set object detection requires models to detect objects of pre-defined categories. (b)  We evaluate models on novel objects and standard Referring expression comprehension (REC) benchmarks for model generalizations on novel objects with attributes. (c) We present an image editing application by combining Grounding DINO and Stable Diffusion \cite{rombach2021highresolution}. Best viewed in colors. 
    }
    \vspace{-0.2cm}
    \label{fig:hero_image}
\end{center}%



\begin{abstract}
In this paper, we develop an open-set object detector, called Grounding DINO, by marrying Transformer-based detector DINO with grounded pre-training, which can detect arbitrary objects with human inputs such as category names or referring expressions. 
The key solution of open-set object detection is introducing language to a closed-set detector for open-set concept generalization. 
To effectively fuse language and vision modalities, we conceptually divide a closed-set detector into three phases and propose a tight fusion solution, which includes a feature enhancer, a language-guided query selection, and a cross-modality decoder for modalities fusion. 
We first pre-train Grounding DINO on large-scale datasets, including object detection data, grounding data, and caption data, and evaluate the model on both open-set object detection and referring object detection benchmarks. 
Grounding DINO performs remarkably well on all three settings, including benchmarks on COCO, LVIS, ODinW, and RefCOCO/+/g. 
Grounding DINO achieves a $52.5$ AP on the COCO zero-shot\footnote{In this paper, ‘zero-shot’ refers to scenarios where the training split of the test dataset is not utilized in the training process.} detection benchmark. It sets a new record on the ODinW zero-shot benchmark with a mean $26.1$ AP.
We release some checkpoints and inference codes at \url{https://github.com/IDEA-Research/GroundingDINO}.
\end{abstract}

\section{Introduction}
\label{sec:intro}

A key indicator of an Artificial General Intelligence (AGI) system's capability is its proficiency in handling open-world scenarios. In this paper, we aim to develop a strong system to detect arbitrary objects specified by human language inputs, a task commonly referred to as \textit{open-set object detection}\footnote{We view the terms \textit{open-set object detection}, \textit{open-world object detection}, and \textit{open-vocabulary object detection} the same task in this paper. To avoid confusion, we always use \textit{open-set object detection} in our paper.}. The task has wide applications for its great potential as a generic object detector. For example, we can cooperate with generative models for image editing (as shown in Fig. \ref{fig:hero_image} (b)). 

{In pursuit of this goal}, we design the strong open-set object detector \textbf{Grounding DINO} by following the two principles: tight modality fusion based on DINO~\cite{zhang2022dino} and large-scale grounded pre-train for concept generalization.

\begin{wrapfigure}[7]{r}{0.7\linewidth}
\centering  
\vspace{-15mm}
\begin{center}
\includegraphics[width=0.8\linewidth]{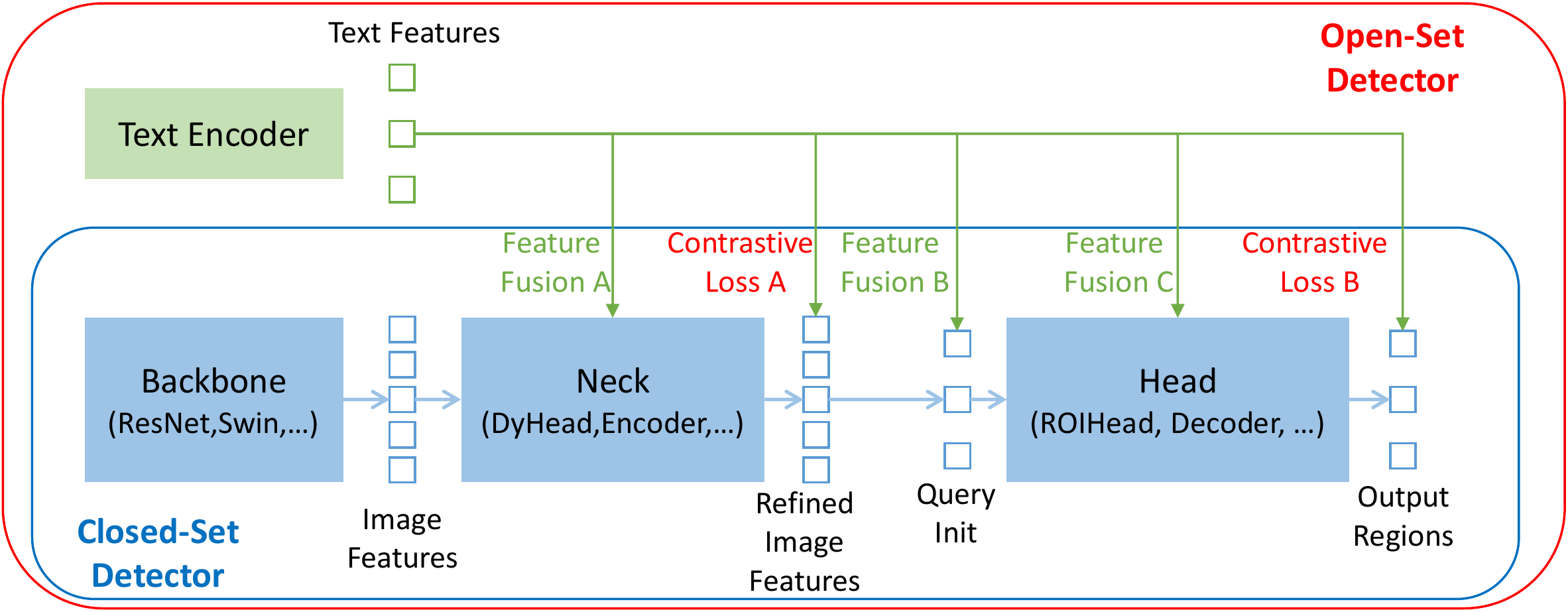}
\end{center}
\caption{Extending closed-set detectors to open-set scenarios. }
\label{fig:model_comparison}  
  \vspace{-6mm}
\end{wrapfigure}

\textbf{Tight modality fusion based on DINO.} 
The key to open-set detection is introducing language for unseen object generalization \cite{li2021grounded, peterAnderson2017BottomUpAT, JiajunDeng2021TransVGEV}. Most existing open-set detectors are developed by extending closed-set detectors to open-set scenarios with language information. As shown in Fig.  \ref{fig:model_comparison}, a closed-set detector typically has three important modules, a backbone for feature extraction, a neck for feature enhancement, and a head for region refinement (or box prediction). A closed-set detector can be generalized to detect novel objects by learning language-aware region embeddings so that each region can be classified into novel categories in a language-aware semantic space. The key to achieving this goal is using contrastive loss between region outputs and language features at the neck and/or head outputs.  

To help a model align cross-modality information 
some work tried to fuse features before the final loss stage. 
We summarize the modulized design of object detectors in Fig.  \ref{fig:model_comparison}.
Feature fusion can be performed in three phases: neck (phase A), query initialization (phase B), and head (phase C). For example, GLIP \cite{li2021grounded} performs early fusion in the neck module (phase A), and OV-DETR \cite{YuhangZang2022OpenVocabularyDW} uses language-aware queries as head inputs (phase B). 
We argue that introducing more feature fusion into the pipeline can facilitate better alignment between different modality features, thereby achieving better performance.


Although conceptually simple, it is hard for previous work to perform feature fusion in all three phases. The design of classical detectors like Faster RCNN makes it hard to interact with language information in most blocks. Unlike classical detectors, the Transformer-based detector {method such as} DINO has a consistent structure with language blocks. The layer-by-layer design enables it to interact with language information easily. 
Under this principle, we design three feature fusion approaches in the neck, query initialization, and head phases. More specifically, we design a feature enhancer by stacking self-attention, text-to-image cross-attention, and image-to-text cross-attention as the neck module. We then develop a language-guided query selection method to initialize queries for the detection head. We also design a cross-modality decoder for the head phase with image and text cross-attention layers to boost query representations. 


\textbf{Large-scale grounded pre-train for zero-shot transfer.}
Most existing open-set models \cite{WeichengKuo2022FindItGL,XiuyeGu2021OpenvocabularyOD} rely on pre-trained CLIP models for concept generalization. 
Nevertheless, the efficacy of CLIP, specifically pre-trained on image-text pairs, is limited for region-text pair detection tasks, as identified in the RegionCLIP study by RegionCLIP \cite{YiwuZhong2022RegionCLIPRL}. 
In contrast, GLIP \cite{li2021grounded} presents a different way by reformulating object detection as a phrase grounding task and introducing contrastive training between object regions and language phrases on large-scale data. It shows great flexibility for heterogeneous datasets and remarkable performance on closed-set and open-set detection. 

We have adopted and refined the grounded training methodology. GLIP's approach involves concatenating all categories into a sentence in a random order. However, the direct category names concatenation does not consider the potential influence of unrelated categories on each other when extracting features. 
To mitigate this issue and improve model performance during grounded training, we introduce a technique that utilizes sub-sentence level text features. It removes the attention between unrelated categories during word feature extractions. Further elaboration on this technique can be found in Section \ref{sec:sub_sentence}.

We pre-train the Grounding DINO on a large-scale dataset and evaluate the performance on mainstream object detection benchmarks like COCO \cite{lin2014microsoft}. 
While some studies have examined open-set detection models under a "partial label" framework—training on a subset of data (e.g., base categories) and testing on additional categories—we advocate for a fully zero-shot approach to enhance practical applicability.
Moreover, we extend the model to another important scenario Referring Expression Comprehension (REC) \cite{Miao2022ReferringEC, JingyuLiu2017ReferringEG}\footnote{We use the term \textit{Referring Expression Comprehension (REC)} and \textit{Referring (Object) Detection} exchangeable in this paper.}, where objects are described with attributes.

We conduct experiments on all three settings, including closed-set detection, open-set detection, and referring object detection, as shown in Fig. \ref{fig:hero_image}, to comprehensively evaluate open-set detection performance. 
Grounding DINO outperforms competitors by a large margin. For example, Grounding DINO reaches a $52.5$ AP on COCO minival without any COCO training data. It also establishes a new state of the art on the ODinW \cite{ChunyuanLi2022ELEVATERAB} zero-shot benchmark with a $26.1$ mean AP.

\begin{table*}[htbp]
\begin{center}
\renewcommand{\arraystretch}{1.1}
\small
\resizebox{\columnwidth}{!}{%
\begin{tabu}{
l@{\hskip9pt} |  
c@{\hskip9pt} c@{\hskip9pt}  c@{\hskip9pt} |
c@{\hskip9pt} |
c@{\hskip9pt} |
c@{\hskip9pt} c@{\hskip9pt} c@{\hskip9pt} |
c@{\hskip9pt} 
}
 \toprule
 \multirow{2}{*}{Model}  & 
 \multicolumn{3}{c|}{Model Design}  & 
 \multicolumn{1}{c|}{Text Prompt}  & 
 \multicolumn{1}{c|}{Closed-Set Settings} &
 \multicolumn{3}{c|}{Zero-Shot Transfer} & 
 \multicolumn{1}{c}{Referring Detection}  
 \\
  & 
  Base Detector & Fusion (Fig. \ref{fig:model_comparison}) &  CLIP & 
  Represent. Level (Sec. \ref{sec:sub_sentence}) & 
  COCO & 
  COCO &  LVIS  & ODinW  &
  RefCOCO/+/g \\
\midrule
  ViLD \cite{XiuyeGu2021OpenvocabularyOD} & Mask R-CNN & - & \checkmark & sentence & \checkmark & partial label & partial label &\\
  RegionCLIP \cite{YiwuZhong2022RegionCLIPRL} & Faster RCNN  & - & \checkmark & sentence & \checkmark & partial label & partial label &\\
  FindIt \cite{WeichengKuo2022FindItGL} & Faster RCNN  & A &  & sentence & \checkmark & partial label & & & fine-tune \\
  MDETR \cite{kamath2021mdetr} & DETR  & A,C &  & word &  & &  fine-tune & zero-shot & fine-tune \\
  DQ-DETR \cite{dqdetr} & DETR  & A,C &  & word & \checkmark & & zero-shot & & fine-tune \\
  GLIP \cite{li2021grounded} & DyHead  & A &  & word & \checkmark & zero-shot & zero-shot & zero-shot & \\
  GLIPv2 \cite{zhang2022glipv2} & DyHead & A &  & word & \checkmark & zero-shot & zero-shot & zero-shot & \\
  OV-DETR \cite{YuhangZang2022OpenVocabularyDW} & Deformable DETR  & B & \checkmark & sentence & \checkmark & partial label & partial label & &  \\
  OWL-ViT \cite{MatthiasMinderer2022SimpleOO} & - & - & \checkmark & sentence & \checkmark & partial label & partial label & zero-shot & \\
  DetCLIP \cite{LeweiYao2022DetCLIPDV} & ATSS  & - & \checkmark & sentence & &  & zero-shot & zero-shot & \\
  OmDet \cite{TianchengZhao2022OmDetLO} & Sparse R-CNN  & C & \checkmark & sentence & \checkmark &  & & zero-shot & \\
  \midrule
  Grounding DINO (Ours) & DINO & A,B,C & & sub-sentence & \checkmark & zero-shot & zero-shot & zero-shot & zero-shot \\
  \bottomrule
\end{tabu}}
\vspace{-0.4cm}
\caption{A comparison of previous open-set object detectors. Our summarization is based on the experiments in their paper, but not the ability to extend their models to other tasks. It is worth noting that some related works may not (only) be designed for the open-set object detection initially, like MDETR \cite{kamath2021mdetr} and GLIPv2\cite{zhang2022glipv2}, but we list them here for a comprehensive comparison with existing work. 
We use the term ``partial label'' for the settings, where models are trained on partial data (e.g. base categories) and evaluated on other cases. \cite{zareian2021open}
\vspace{-0.2cm}
}
\label{table:related_work}
\end{center}
\end{table*}
\section{Related Work}
\label{sec:related_work}

\textbf{Detection Transformers.}
Grounding DINO is built upon the DETR-like model DINO \cite{zhang2022dino}, which is an end-to-end Transformer-based detector. 
DETR was first proposed in \cite{carion2020end} and then has been improved from many directions \cite{zhu2020deformable, meng2021conditional, gao2021fast, dai2021dynamic, anchordetr, HDETR, groupdetr} in the past few years.
DAB-DETR \cite{liu2022dabdetr} introduces anchor boxes as DETR queries for more accurate box prediction. DN-DETR \cite{li2022dn} proposes a query-denoising approach to stabilizing the bipartite matching. 
DINO \cite{zhang2022dino} further develops several techniques including contrastive de-noising and sets a new record on the COCO object detection benchmark.
However, such detectors mainly focus on closed-set detection and are difficult to generalize to novel classes because of the limited pre-defined categories.

\noindent
\textbf{Open-Set Object Detection.} Open-set object detection is trained using existing bounding box annotations and aims at detecting arbitrary classes with the help of language generalization. OV-DETR \cite{zareian2021open} uses image and text embedding encoded by a CLIP model as queries to decode the category-specified boxes in the DETR framework \cite{carion2020end}. ViLD \cite{XiuyeGu2021OpenvocabularyOD} distills knowledge from a CLIP teacher model into a R-CNN-like detector so that the learned region embeddings contain the semantics of language. GLIP \cite{gao2021clip} formulates object detection as a grounding problem and leverages additional grounding data to help learn aligned semantics at phrase and region levels. It shows that such a formulation can even achieve stronger performance on fully-supervised detection benchmarks. DetCLIP \cite{LeweiYao2022DetCLIPDV} involves large-scale image captioning datasets and uses the generated pseudo labels to expand the knowledge database. The generated pseudo labels effectively help extend the generalization ability.

However, previous works only fuse multi-modal information in partial phases, which may lead to sub-optimal language generalization ability. 
For example, GLIP only considers fusion in the feature enhancement (phase A) and OV-DETR only injects language information at the decoder inputs (phase B).
Moreover, the REC task is normally overlooked in evaluation, which is an important scenario for open-set detection. We compare our model with other open-set methods in Table \ref{table:related_work}.

\begin{figure*}[t]
    \centering
    \includegraphics[width=1.0\linewidth]{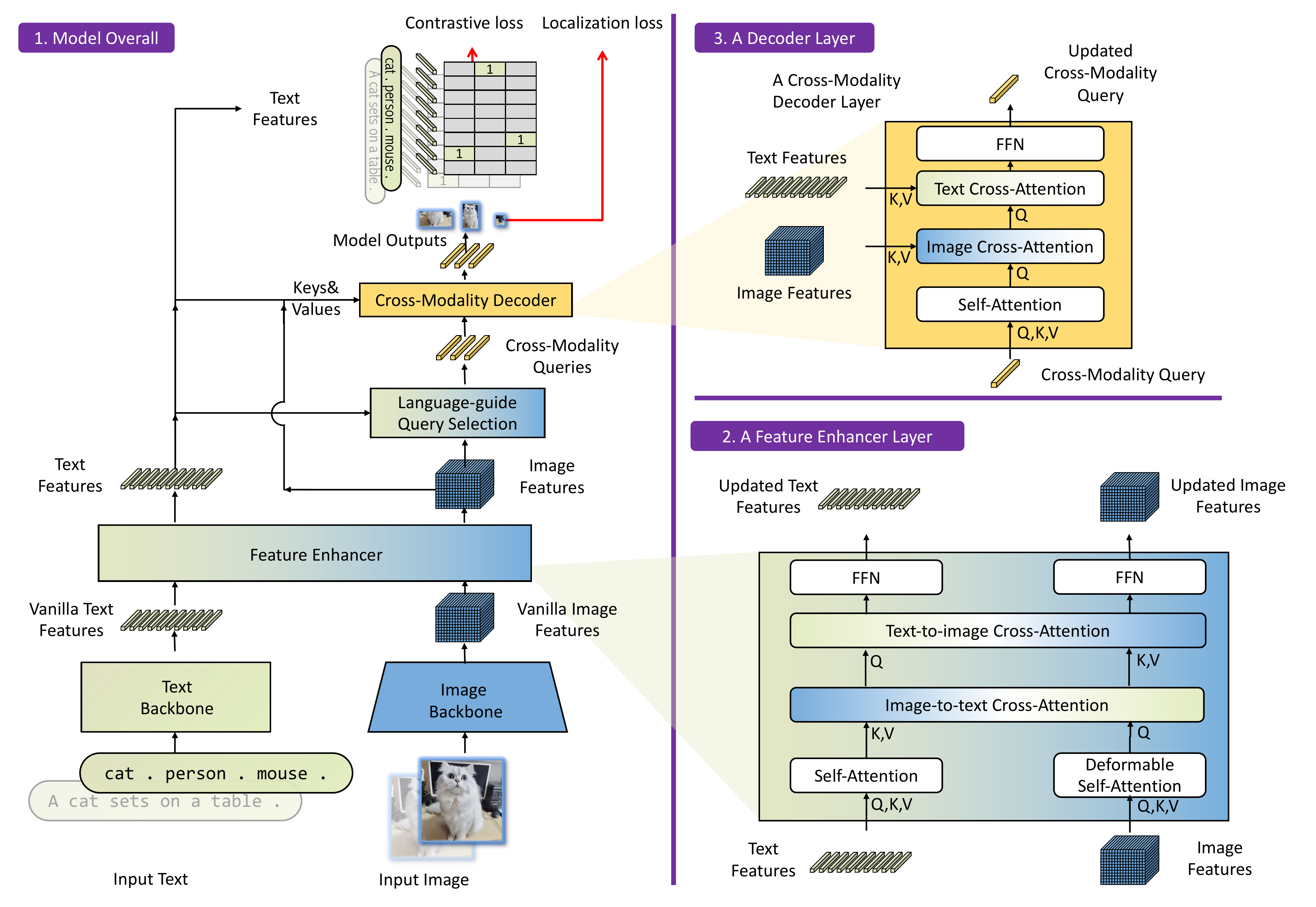}
    \caption{The framework of Grounding DINO. We present the overall framework, a feature enhancer layer, and a decoder layer in block 1, block 2, and block 3, respectively. }
    \label{fig:framework}
\end{figure*}
\section{Grounding DINO}
\label{sec:groundingdino}

Grounding DINO outputs multiple pairs of object boxes and noun phrases for a given \texttt{(Image, Text)} pair. For example, as shown in Fig. \ref{fig:framework}, the model locates a cat and a table from the input image and extracts word \texttt{cat} and \texttt{table} from the input text as corresponding labels. Both object detection and REC tasks can be aligned with the pipeline. Following GLIP \cite{li2021grounded}, we concatenate all category names as input texts for object detection tasks. REC requires a bounding box for each text input. We use the output object with the largest scores as the output for the REC task.

Grounding DINO is a dual-encoder-single-decoder architecture. It contains an image backbone for image feature extraction, a text backbone for text feature extraction, a feature enhancer for image and text feature fusion (Sec. \ref{sec:feature_enhance}), a language-guided query selection module for query initialization (Sec. \ref{sec:query_selection}), and a cross-modality decoder for box refinement (Sec. \ref{sec:cross_modal_decoder}). 

For each \texttt{(Image, Text)} pair, we first extract vanilla image features and vanilla text features using an image backbone and a text backbone, respectively. 
The two vanilla features are fed into a feature enhancer module for cross-modality feature fusion. After obtaining cross-modality text and image features, we use a language-guided query selection module to select cross-modality queries from image features. Like the object queries in most DETR-like models, these cross-modality queries will be fed into a cross-modality decoder to probe desired features from the two modal features and update themselves. The output queries of the last decoder layer will be used to predict object boxes and extract corresponding phrases.

\begin{figure}
    \vspace{-2mm}
    \centering
    \includegraphics[width=0.7\linewidth]{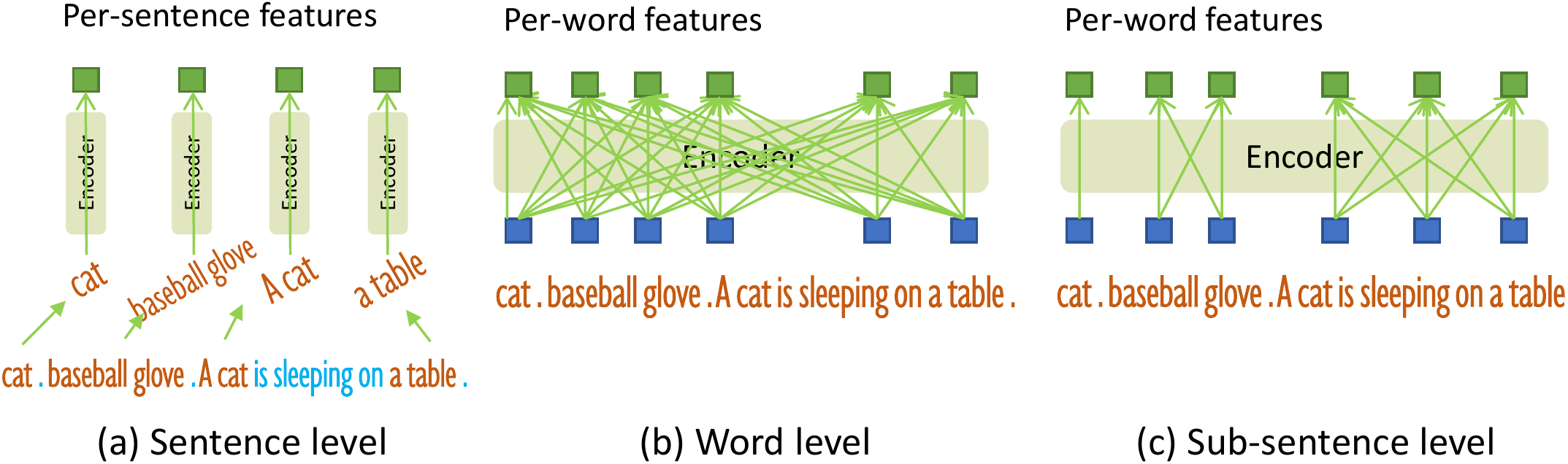}
    \caption{Comparisons of text representations. }
    \label{fig:text_prompts}
    \vspace{-2mm}
\end{figure}

\vspace{-2mm}
\subsection{Feature Extraction and Enhancer}
\label{sec:feature_enhance}
Given an \texttt{(Image, Text)} pair, we extract multi-scale image features with an image backbone like Swin Transformer \cite{liu2021swin}, and text features with a text backbone like BERT \cite{devlin2018bert}. Following previous DETR-like detectors \cite{zhu2020deformable, zhang2022dino}, multi-scale features are extracted from the outputs of different blocks. 
After extracting vanilla image and text features, we fed them into a feature enhancer for cross-modality feature fusion. 
The feature enhancer includes multiple feature enhancer layers.
We illustrate a feature enhancer layer in Fig. \ref{fig:framework} block 2. 
We leverage the Deformable self-attention to enhance image features and the vanilla self-attention for text feature enhancers. 
Inspired by GLIP \cite{li2021grounded}, we add an image-to-text and a text-to-image cross-attention modules for feature fusion.
These modules help align features of different modalities.

\subsection{Language-Guided Query Selection}
\label{sec:query_selection}
Grounding DINO aims to detect objects from an image specified by an input text. To effectively leverage the input text to guide object detection, we design a language-guided query selection module to select features that are more relevant to the input text as decoder queries. 




Let's denote the image feature as $\Xmat_{I} \in \textsc{R}^{N_I\times d}$ and the text features as $\Xmat_{T}\in \textsc{R}^{N_T\times d}$. Here, $N_I$ represents the number of image tokens, $N_T$ indicates the number of text tokens, and $d$ corresponds to the feature dimension. In our experiments, we specifically utilize a feature dimension of $d=256$. Typically, in our models, the value of $N_I$ exceeds $10,000$, while $N_T$ remains below $256$.
Our objective is to extract $N_q$ queries from the encoder's image features to be used as inputs for the decoder. In alignment with the DINO method, we set $N_q$ to be $900$. The top $N_q$ query indices for the image feature, denoted as $\Imat_{N_q}$, are selected using the following expression:

\begin{equation}
\Imat_{N_q} = \texttt{Top}_{{N_q}}(\texttt{Max}^{(-1)}(\Xmat_{I}\Xmat_{T}^{\intercal})).
\end{equation}

In this expression, $\texttt{Top}_{{N_q}}$ represents the operation to pick the top $N_q$ indices. The function $\texttt{Max}^{(-1)}$ executes the \texttt{max} operation along the $-1$ dimension, and the symbol $^{\intercal}$ denotes matrix transposition.
We present the query selection process in Algorithm \ref{code:qs} in PyTorch style. 
The language-guided query selection module outputs $N_q$ indices. We can extract features based on the selected indices to initialize queries. Following DINO \cite{zhang2022dino}, we use mixed query selection to initialize decoder queries. Each decoder query contains two parts: content part and positional part \cite{meng2021conditional}, respectively. We formulate the positional part as dynamic anchor boxes \cite{liu2022dabdetr}, which are initialized with encoder outputs. The other part, the content queries, are set to be learnable during training.

\subsection{Cross-Modality Decoder}
\label{sec:cross_modal_decoder}
We develop a cross-modality decoder to combine image and text modality features, as shown in Fig. \ref{fig:framework} block 3. Each cross-modality query is fed into a self-attention layer, an image cross-attention layer to combine image features, a text cross-attention layer to combine text features, and an FFN layer in each cross-modality decoder layer. 
Each decoder layer has an extra text cross-attention layer compared with the DINO decoder layer, as we need to inject text information into queries for better modality alignment.

\subsection{Sub-Sentence Level Text Feature}
\label{sec:sub_sentence}

Two kinds of text prompts are explored in previous works, which we named as sentence level representation and word level representation, as shown in Fig.~\ref{fig:text_prompts}. Sentence level representation \cite{LeweiYao2022DetCLIPDV,MatthiasMinderer2022SimpleOO} encodes 
a whole sentence to one feature. 
If some sentences in phrase grounding data have multiple phrases, it extracts these phrases and discards other words.
In this way, it removes the influence between words while losing fine-grained information in sentences. 
Word level representation \cite{gao2021clip,kamath2021mdetr} enables encoding multiple category names with one forward but introduces unnecessary dependencies among categories, especially when the input text is a concatenation of multiple category names in an arbitrary order. As shown in Fig. \ref{fig:text_prompts} (b), some unrelated words interact during attention. To avoid unwanted word interactions, we introduce attention masks to block attentions among unrelated category names, named ``sub-sentence'' level representation. It eliminates the influence between different category names while keeping per-word features for fine-grained understanding.

\vspace{-0.2cm}

\subsection{Loss Function}

Following previous DETR-like works \cite{carion2020end, zhu2020deformable, meng2021conditional, liu2022dabdetr, li2022dn, zhang2022dino}, we use the L1 loss and the GIOU \cite{rezatofighi2019generalized} loss for bounding box regressions. We follow GLIP \cite{li2021grounded} and use contrastive loss between predicted objects and language tokens for classification.
Specifically, we dot product each query with text features to predict logits for each text token and then compute focal loss \cite{lin2017focal} for each logit. 
Box regression and classification costs are first used for bipartite matching between predictions and ground truths. We then calculate final losses between ground truths and matched predictions with the same loss components.
Following DETR-like models, we add auxiliary loss after each decoder layer and after the encoder outputs.

\begin{table*}[t]
\caption{Zero-shot domain transfer and fine-tuning on COCO. * The results in brackets are trained with 1.5$\times$ image sizes, i.e., with a maximum image size of 2000. \dag The models map a subset of O365 categories to COCO for zero-shot evaluations. \ddag This is not a real zero-shot performance as we add COCO data during model training. We list the result here for a reference.}
\label{table:cocomain}
\begin{center}
\resizebox{\linewidth}{!}{
\begin{tabular}{l@{\hskip9pt} 
c@{\hskip9pt} | c | c 
c  c@{\hskip9pt}c@{\hskip9pt}
 }
\toprule
\multirow{2}{*}{Model} & 
\multirow{2}{*}{Backbone} & 
\multirow{2}{*}{Pre-Training Data}  &
{Zero-Shot}  & Fine-Tuning   \\
 & 
 & 
 &
\small{2017val}  & \small{2017val/test-dev}   \\
\midrule
Faster R-CNN & RN50-FPN & -  & - & 40.2 / - \\
Faster R-CNN & RN101-FPN & -  & - & 42.0 / - \\
DyHead-T \cite{dai2021dynamic} & Swin-T & - & - & 49.7 / - \\
DyHead-L \cite{dai2021dynamic} & Swin-L & - & - & 58.4 / 58.7 \\
DyHead-L \cite{dai2021dynamic} & Swin-L  & O365,ImageNet21K  & - & 60.3 / 60.6\\
 SoftTeacher \cite{xu2021end} & Swin-L  & O365,SS-COCO & - & 60.7 / 61.3 \\
 DINO(Swin-L) \cite{zhang2022dino} & Swin-L  & O365 & - & 62.5 / - \\
\midrule
DyHead-T\dag \cite{dai2021dynamic} & Swin-T &  O365   & 43.6 & 53.3 / - \\
GLIP-T (B) \cite{li2021grounded} & Swin-T     & O365  & 44.9 & 53.8 / - \\
GLIP-T (C)  \cite{li2021grounded} & Swin-T     & O365,GoldG & {46.7} & 55.1 / - \\
GLIP-L  \cite{li2021grounded} & Swin-L  & FourODs,GoldG,Cap24M & {49.8} & {60.8} / 61.0 \\
\midrule
DINO(Swin-T)\dag \cite{zhang2022dino} & Swin-T & O365 & 46.2  & 56.9 / - \\
\rowcolor{Gray}
Grounding DINO T (Ours)   & Swin-T    & O365    & 46.7  & 56.9 / -  \\
\rowcolor{Gray}
Grounding DINO T (Ours)  & Swin-T    & O365,GoldG    & 48.1  & 57.1 / -  \\
\rowcolor{Gray}
Grounding DINO T (Ours)  & Swin-T    & O365,GoldG,Cap4M    & 48.4  & 57.2 / -  \\
\rowcolor{Gray}
Grounding DINO L (Ours)   & Swin-L    & O365,OI\cite{krasin2017openimages},GoldG    & \textbf{52.5}  & \textbf{62.6} / \textbf{62.7} (\textbf{63.0} / \textbf{63.0})* \\
\midrule
\rowcolor{Gray}
Grounding DINO L (Ours)  & Swin-L    & O365,OI,GoldG,Cap4M,COCO,RefC    & \textbf{60.7}\ddag  & \textbf{62.6} / -\\
\bottomrule
\end{tabular}
}
\end{center}
\vspace{-6mm}
\end{table*}


\section{Experiments}
\label{sec:exp}

We conduct extensive experiments on three settings: a closed-set setting on the COCO detection benchmark (Sec. \ref{sec:coco}), an open-set setting on zero-shot COCO, LVIS, and ODinW (Sec. \ref{sec:open-set}), and a referring detection setting on RefCOCO/+/g (Sec. \ref{sec:visual_grounding}). 
Ablations are then conducted to show the effectiveness of our model design (Sec. \ref{sec:ablations}).
We also explore a way to transfer a well-trained DINO to the open-set scenario by training a few plug-in modules in Sec. \ref{sec:dino_to_groundingdino}.
The test of our model efficiency is presented in Sec. \ref{sec:efficiency}.

\subsection{Implementation Details}

We trained two model variants, Grounding DINO T with Swin-T \cite{liu2021swin}, and Grounding DINO L with Swin-L \cite{liu2021swin} as an image backbone, respectively. We leveraged BERT-base \cite{devlin2018bert} from Hugging Face \cite{wolf2019huggingface} as text backbones. 
As we focus more on the model performance on novel classes, we list zero-shot transfer and referring detection results in the main text. 

By default, we use 900 queries in our model following DINO. We set the maximum text token number as 256. Using BERT as our text encoder, we follow BERT to tokenize texts with a BPE scheme \cite{bpescheme}. We use six feature enhancer layers in the feature enhancer module. The cross-modality decoder is composed of six decoder layers as well. 
We leverage deformable attention \cite{zhu2020deformable} in image cross-attention layers.

Both matching costs and final losses include classification losses (or contrastive losses), box L1 losses, and GIOU \cite{rezatofighi2019generalized} losses. Following DINO, we set the weight of classification costs, box L1 costs, and GIOU costs as 2.0, 5.0, and 2.0, respectively, during Hungarian matching. The corresponding loss weights are 1.0, 5.0, and 2.0 in the final loss calculation.

Our Swin Transformer Tiny models are trained on 16 Nvidia V100 GPUs with a total batch size of 32. We extract three image feature scales, from 8$\times$ to 32$\times$. It is named ``4scale'' in DINO since we downsample the 32$\times$ feature map to 64$\times$ as an extra feature scale. 
For the model with Swin Transformer Large, we extract four image feature scales from backbones, from 4$\times$ to 32$\times$. The model is trained on 64 Nvidia A100 GPUs with a total batch size of 64.

\subsection{Zero-Shot Transfer of Grounding DINO}
\label{sec:open-set}


In this setting, we pre-train models on large-scale datasets and directly evaluate models on new datasets. We also list some fine-tuned results for a more thorough comparison of our model with prior works.


\paragraph{COCO Benchmark}
\label{sec:coco_exp}

We compare Grounding DINO with GLIP and DINO in Table \ref{table:cocomain}. 
We pre-train models on large-scale datasets and directly evaluate our model on the COCO benchmark. As the O365 dataset \cite{shao2019objects365} has (nearly\footnote{It is not an exact mapping between O365 and COCO categories. We made some approximations during evaluation. }) covered all categories in COCO, we evaluate an O365 pre-trined DINO on COCO as a zero-shot baseline. The result shows that DINO performs better on the COCO zero-shot transfer than DyHead. 
Grounding DINO outperforms all previous models on the zero-shot transfer setting, with $+0.5$AP and $+1.8$AP compared with DINO and GLIP under the same setting. Grounding data is still helpful for Grounding DINO, introducing more than $1$AP (48.1 vs. 46.7) on the zero-shot transfer setting. 
With stronger backbones and larger data, Grounding DINO sets a new record of $52.5$ AP on the COCO object detection benchmark without seeing any COCO images during training.
Grounding DINO obtains a $62.6$ AP on COCO minival, outperforming DINO's $62.5$ AP. 
When enlarging the input images by $1.5 \times$, the benefits reduce. We suspect that the text branch enlarges the gap between models with different input images. {Even though the performance plateaus with larger input size}, Grounding DINO gets an impressive $63.0$ AP on COCO test-dev with fine-tuning on the COCO dataset(See the number in brackets of Table \ref{table:cocomain}).


\paragraph{LVIS Benchmark}
\label{sec:lvis_exp}

LVIS \cite{gupta2019lvis} is a dataset for long-tail objects. It contains more than $1000$ categories for evaluation. We use LVIS as a downstream task to test the zero-shot abilities of our model. We use GLIP and DetCLIPv2 as baselines for our models. The results are shown in Table \ref{table:zslvis}. 

\begin{table}
\resizebox{0.7\linewidth}{!}{
\begin{tabular}{l@{\hskip9pt} l@{\hskip9pt} |
c@{\hskip9pt} | c@{\hskip9pt}c@{\hskip9pt} }
\toprule

\multirow{2}{*}{Model} & 
\multirow{2}{*}{Backbone} &
\multirow{2}{*}{Pre-Training Data} &
\multicolumn{2}{c}{MiniVal \cite{kamath2021mdetr}} \\
 && & AP & APr/APc/APf  \\
\midrule
\multicolumn{5}{c}{\textit{Zero-Shot Setting}} \\
\midrule
GLIP-T (C) & Swin-T & O365,GoldG  & 24.9 & 17.7/19.5/{31.0}    \\
GLIP-T & Swin-T & O365,GoldG,Cap4M &  26.0 & {20.8}/21.4/31.0  \\
DetCLIPv2 & Swin-T & O365,GoldG,CC15M & 40.4 & 36.0/41.7/40.0 \\
\midrule
\rowcolor{Gray} Grounding DINO T  & Swin-T & O365,GoldG  & 25.6 & 14.4/19.6/32.2 \\
\rowcolor{Gray} Grounding DINO T  & Swin-T  & O365,GoldG,Cap4M & {{27.4}} & 18.1/{23.3}/{32.7} \\
\rowcolor{Gray} Grounding DINO L & Swin-L & \makecell{O365,OI,GoldG,Cap4M,\\COCO,RefC} & 33.9 & 22.2/30.7/38.8\\
\midrule
\multicolumn{5}{c}{\textit{Fine-Tune Setting}} \\
\midrule
 MDETR &  RN101 & GoldG,RefC & 24.2 &  20.9/24.9/24.3    \\
 Mask R-CNN &  RN101 & -  &  33.3 &  26.3/34.0/33.9  \\
DetCLIPv2 \cite{yao2023detclipv2} & Swin-T & O365,GoldG,CC15M  & 50.7 & 44.3/52.4/{50.3}    \\
\midrule
\rowcolor{Gray} Grounding DINO T  & Swin-T & O365,GoldG  & \textbf{52.1} & 35.4/51.3/55.7 \\
\bottomrule
\end{tabular}
}
\caption{Model results on LVIS. }
\label{table:zslvis}
\end{table}
\vspace{-0.4cm}

We found two interesting phenomena in the results. First, Grounding DINO works better than common objects than GLIP, but worse on rare categories. 
We reviewed DETR-like models on LVIS and noted these models often exhibit lower rare category AP despite similar overall AP, like Table 2 of \cite{dong2023boostinglongtailedobjectdetection} and Table 6 of \cite{kamath2021mdetr}. To our knowledge, no existing DETR-like models effectively address the rarity challenge in LVIS without extra training data, which may be a characteristic limitation of the architecture.

The other phenomenon is that Grounding DINO has larger gains with more data than GLIP. For example, Grounding DINO introduces $+1.8$ AP gains with the caption data Cap4M, whereas GLIP has only $+1.1$ AP. We believe that Grounding DINO has better scalability compared with GLIP. 
A larger-scale training will be left as our future work.

Although achieving better results than GLIP, we found that Grounding DINO is inferior to DetCLIPv2, which is trained on a larger scale data. This performance difference might be attributed to the disparity in data distribution between the training dataset and the LVIS dataset.

To unveil the full potential of Grounding DINO, we fine-tuned it on the LVIS dataset. Table \ref{table:zslvis} highlights the commendable capability of our model. Remarkably, despite being pre-trained only on the O365 and GoldG datasets, Grounding DINO outperforms DetCLIPv2-T by a margin of $1.5$ AP. This result shows that Grounding DINO might have learned a better object-level representation which helps yield a better performance after fine-tuning (aligning with the target dataset). In our future work, we will perform more studies, including varying the semantic concept coverage of the training data and increasing the scale of the training data, to further improve the zero-shot generalization performance.


\begin{table*}[htbp]
\caption{Model results on the ODinW benchmark. }
\label{tab:odinw}
\begin{center}
\small
\resizebox{\linewidth}{!}{
\begin{tabular}{l@{\hskip9pt}c@{\hskip9pt} | c@{\hskip9pt}c@{\hskip9pt} | c@{\hskip9pt} | cc}
 \toprule
 \multirow{2}{*}{Model}  & 
 \multirow{2}{*}{Language Input}  & 
 \multirow{2}{*}{Backbone}  &  
 \multirow{2}{*}{Model Size}  & 
 \multirow{2}{*}{Pre-Training Data}  & 
 \multicolumn{2}{c}{Test} \\
  & & & &  & AP$_{average}$ & AP$_{median}$\\
 \midrule
   \multicolumn{7}{c}{\textit{Zero-Shot Setting}}  \\
 \midrule
 MDETR \cite{kamath2021mdetr} & \ding{51} & ENB5 \cite{MingxingTan2019EfficientNetRM} & 169M & GoldG,RefC & 10.7 & 3.0 \\
 OWL-ViT \cite{MatthiasMinderer2022SimpleOO} & \ding{51} & ViT L/14(CLIP) & $>$1243M & O365, VG & 18.8 & {9.8} \\
 GLIP-T \cite{li2021grounded} & \ding{51} & Swin-T & 232M & O365,GoldG,Cap4M & 19.6 & 5.1 \\
 OmDet \cite{TianchengZhao2022OmDetLO} & \ding{51} & ConvNeXt-B & 230M & COCO,O365,LVIS,PhraseCut & 19.7 & {10.8} \\
 GLIPv2-T \cite{GLIPv2} & \ding{51} & Swin-T & 232M & O365,GoldG,Cap4M & {22.3} & 8.9 \\
 DetCLIP \cite{LeweiYao2022DetCLIPDV} & \ding{51} & Swin-L & 267M & O365,GoldG,YFCC1M & 24.9	 & 18.3 \\
 Florence  \cite{LuYuan2022FlorenceAN} & \ding{51} & CoSwinH & $\approx$841M & FLD900M,O365,GoldG & 25.8	 & 14.3 \\
 \midrule
 \rowcolor{Gray} Grounding DINO T(Ours) & \ding{51} & Swin-T & 172M & O365,GoldG & {20.0} & 9.5 \\
 \rowcolor{Gray} Grounding DINO T(Ours) & \ding{51} & Swin-T & 172M & O365,GoldG,Cap4M & {22.3} & {11.9} \\
 \rowcolor{Gray} Grounding DINO L(Ours) & \ding{51} & Swin-L & 341M & O365,OI,GoldG,Cap4M,COCO,RefC & \textbf{26.1} & \textbf{18.4} \\
 \midrule
    \multicolumn{7}{c}{\textit{Few-Shot Setting}}  \\
 \midrule
 DyHead-T \cite{dai2021dynamic} & \ding{55} & Swin-T & $\approx$100M & O365 & 37.5 & 36.7 \\
 GLIP-T \cite{li2021grounded} & \ding{51} & Swin-T & 232M & O365,GoldG,Cap4M & 38.9 & 33.7 \\
 DINO-Swin-T \cite{zhang2022dino} &  \ding{55} & Swin-T & 49M & O365 & 41.2 & 41.1 \\
 OmDet \cite{TianchengZhao2022OmDetLO} & \ding{51} & ConvNeXt-B & 230M & COCO,O365,LVIS,PhraseCut & 42.4 & 41.7 \\
 \midrule
 \rowcolor{Gray}
 Grounding DINO T(Ours) & \ding{51} & Swin-T & 172M & O365,GoldG & \textbf{46.4} & \textbf{51.1} \\
  \midrule
    \multicolumn{7}{c}{\textit{Full-Shot Setting}}  \\
 \midrule
 GLIP-T \cite{li2021grounded} & \ding{51} & Swin-T & 232M & O365,GoldG,Cap4M & 62.6 & 62.1 \\
 DyHead-T \cite{dai2021dynamic} & \ding{55} & Swin-T & $\approx$100M & O365 & 63.2 & 64.9 \\
 DINO-Swin-T \cite{zhang2022dino} & \ding{55} & Swin-T & 49M & O365 & 66.7 & 68.5 \\
 OmDet \cite{TianchengZhao2022OmDetLO} & \ding{51} & ConvNeXt-B & 230M & COCO,O365,LVIS,PhraseCut & 67.1 & 71.2 \\
 DINO-Swin-L \cite{zhang2022dino} & \ding{55} & Swin-L & 218M & O365 & 68.8 & 70.7 \\
 \midrule
 \rowcolor{Gray}
 Grounding DINO T(Ours) & \ding{51}  & Swin-T & 172M & O365,GoldG & \textbf{70.7} & \textbf{76.2} \\
\bottomrule
\end{tabular}
}
\vspace{-4mm}
\end{center}
\end{table*}

\vspace{-0.7cm}

\paragraph{ODinW Benchmark}
\label{sec:odinw_exp}

ODinW (Object Detection in the Wild) \cite{ChunyuanLi2022ELEVATERAB} is a more challenging benchmark to test model performance under real-world scenarios. It collects more than $35$ datasets for evaluation. We report three settings, zero-shot, few-shot, and full-shot results in Table \ref{tab:odinw}. Grounding DINO performs well on this benchmark. With only O365 and GoldG for pre-train, Grounding DINO T outperforms DINO on few-shot and full-shot settings. Impressively, Grounding DINO with a Swin-T backbone outperforms DINO with Swin-L on the full-shot setting. 

Grounding DINO outperforms GLIP under the same backbone for the zero-shot setting.
{Grounding DINO and GLIPv2-T show similar $AP_{average}$. However, a key distinction lies in the $AP_{median}$, where Grounding DINO significantly outperforms GLIPv2-T (11.9 vs 8.9). This suggests that while GLIPv2 may exhibit larger performance variance across different datasets, Grounding DINO maintains a more consistent performance level. GLIPv2 incorporates advanced techniques like masked text training and cross-instance contrastive learning, making it more complex than our Grounding DINO model. Moreover, our model is more compact (172M parameters) compared to GLIPv2 (232M parameters). These factors combined—performance consistency, model complexity, and size—should address concerns about our model's capability in true open-set scenarios.}

Grounding DINO L set a new record on ODinW zero-shot with a $26.1$ AP, even outperforming the giant Florence models \cite{LuYuan2022FlorenceAN}. The results show the generalization and scalability of Grounding DINO.




\subsection{Referring Object Detection Settings}
\label{sec:visual_grounding}


\label{sec:refcoco_exp}
We further explore our models' performances on the REC task. 
We leverage GLIP \cite{li2021grounded} as our baseline. We evaluate the model performance on RefCOCO/+/g directly.\footnote{We used the official released code and checkpoints in \url{https://github.com/microsoft/GLIP}.} 
The results are shown in Table \ref{table:refexp}. Grounding DINO outperforms GLIP under the same setting. Nevertheless, both GLIP and Grounding DINO perform not well without REC data. More training data like caption data or larger models help the final performance, but quite minor. After injecting RefCOCO/+/g data into training, Grounding DINO obtains significant gains. The results reveal that most nowadays open-set object detectors need to pay more attention for a more fine-grained detection. 

\begin{table}[!htbp]
\renewcommand{\arraystretch}{1.1}
\centering
\resizebox{\linewidth}{!}{%
 \begin{tabular}{l|c|c|c|ccc|ccc|cc} 
 \toprule
 \multirow{2}{*}{Method}  & 
  \multirow{2}{*}{Backbone}  & 
  \multirow{2}{*}{Pre-Training Data} & 
 \multirow{2}{*}{Fine-tuning} & 
 \multicolumn{3}{c|}{RefCOCO} &  
 \multicolumn{3}{c|}{RefCOCO+} & 
 \multicolumn{2}{c}{RefCOCOg}  \\ [0.5ex] 
         & & & & val & testA & testB & val & testA & testB & val & test  \\
 \midrule
 MAttNet~\cite{LichengYu2018MAttNetMA} & R101 & None & \checkmark &76.65  & 81.14 & 69.99 & 65.33 & 71.62 & 56.02 & 66.58 & 67.27  \\
 VGTR~\cite{YeDu2021VisualGW} & R101& None & \checkmark & 79.20 & 82.32 & 73.78 & 63.91 & 70.09 & 56.51 & 65.73 & 67.23 \\
 TransVG~\cite{JiajunDeng2021TransVGEV}  & R101& None & \checkmark & 81.02 & 82.72 & 78.35 & 64.82 & 70.70 &  56.94 & 68.67 & 67.73 \\
 VILLA\_L$^*$~\cite{ZheGan2020LargeScaleAT}  & R101& CC, SBU, COCO, VG & \checkmark & 82.39 & 87.48  & 74.84 &  76.17 & 81.54 &  66.84 & 76.18 & 76.71   \\ 
 RefTR~\cite{MuchenLi2021ReferringTA} & R101 & VG  & \checkmark & 85.65 & 88.73 & 81.16 & 77.55 & 82.26 & 68.99 & 79.25 & 80.01 \\
 MDETR~\cite{kamath2021mdetr}  & R101& GoldG,RefC  & \checkmark & {86.75} & {89.58}& {81.41}     & {79.52}  & {84.09} &  {70.62}  & {81.64} & {80.89}  \\
 DQ-DETR \cite{dqdetr} & R101& GoldG,RefC & \checkmark & {88.63} & {91.04} & {83.51}    & {{81.66}}  & {86.15} &  {73.21}  & {82.76} & {83.44}  \\
\midrule
 GLIP-T(B) & Swin-T & O365,GoldG & & 49.96 & 54.69 & 43.06 & 49.01 & 53.44 & 43.42 & 65.58 & 66.08 \\
 GLIP-T & Swin-T & O365,GoldG,Cap4M & & 50.42 & 54.30 & 43.83 & 49.50 & 52.78 & 44.59 & 66.09 & 66.89 \\
\midrule
 \rowcolor{Gray} Grounding DINO T (Ours) & Swin-T & O365,GoldG            &  & 50.41 & 57.24 & 43.21 & 51.40 & 57.59 & 45.81 & 67.46 & 67.13 \\ 
 \rowcolor{Gray} Grounding DINO T (Ours) & Swin-T & O365,GoldG,RefC            &  & 73.98 & 74.88 & 59.29 & 66.81 & 69.91 & 56.09  & 71.06 & 72.07 \\
 \rowcolor{Gray} Grounding DINO T (Ours) & Swin-T & O365,GoldG,RefC            & \checkmark & {89.19} & {91.86} & {85.99} & 81.09 & {87.40} & {74.71} & {84.15} & {84.94} \\
  \midrule
  \rowcolor{Gray} Grounding DINO L (Ours)* & Swin-L & O365,OI,GoldG,Cap4M,COCO,RefC & \checkmark & \textbf{90.56} & \textbf{93.19} & \textbf{88.24} & \textbf{82.75} & \textbf{88.95} & \textbf{75.92} & \textbf{86.13} & \textbf{87.02} \\
\bottomrule
\end{tabular}
}
\caption{
\small{
Top-1 accuracy comparison on the referring expression comprehension task. 
We mark the best results in bold.
All models are trained with a ResNet-101 backbone. 
We use the notations ``CC'', ``SBU'', ``VG'', ``OI'', ``O365'', and ``YFCC'' for Conceptual Captions~\cite{PiyushSharma2018ConceptualCA}, SBU Captions~\cite{VicenteOrdonez2011Im2TextDI}, Visual Genome~\cite{RanjayKrishna2017VisualGC}, OpenImage~\cite{AlinaKuznetsova2018TheOI}, Objects365~\cite{zhou2019objects}, YFCC100M~\cite{BartThomee2016YFCC100MTN} respectively.  The term ``RefC'' is used for RefCOCO, RefCOCO+, and RefCOCOg three datasets.
* There might be a data leak since COCO includes validation images in RefC. But the annotations of the two datasets are different.
} 
}
\label{table:refexp}
\vspace{-2mm}
\end{table}
\vspace{-0.2cm}

\subsection{Effects of RefC and COCO Data}

We add the RefCOCO/+/g (we note it as ``RefC'' in tables) and COCO into training in some settings. We explore the influence of these data in Table \ref{table:add_ref}. The results show that RefC helps improve the COCO zero-shot and fine-tuning performance but hurts the LVIS and ODinW results. With COCO introduced, the COCO results are greatly improved. It shows that COCO brings marginal improvements in LVIS and slight decreases on ODinW.
\begin{table}[ht]
\begin{center}
\small
\resizebox{0.7\columnwidth}{!}{%
\begin{tabu}{
l@{\hskip9pt} |  
c@{\hskip9pt} |  
c@{\hskip9pt} c@{\hskip9pt}  |
c@{\hskip9pt}  | 
c@{\hskip9pt} 
}
 \toprule
 \multirow{2}{*}{Model}  & 
 \multirow{2}{*}{Pre-Train}  & 
 \multicolumn{2}{c|}{COCO minival} &
 \multicolumn{1}{c|}{LVIS minival} &
 \multicolumn{1}{c}{ODinW} 
 \\
  &
  &
  \small{Zero-Shot}  & \small{Fine-Tune} &
  \small{Zero-Shot} &
  \small{Zero-Shot}
  \\
\midrule
  Grounding DINO T  & O365,GoldG  & 48.1 & 57.1 & 25.6 & 20.0  \\
  Grounding DINO T  & O365,GoldG,RefC  & 48.5 & 57.3 & 21.9 & 17.7  \\
  Grounding DINO T  & O365,GoldG,RefC,COCO  & 56.1 & 57.5  & 22.3 & 17.4   \\
  \bottomrule
\end{tabu}}
\vspace{-0.5cm}
\caption{Impacts of RefC and COCO data for open-set settings. All models are trained with a Swin Transformer Tiny backbone.}
\label{table:add_ref}
\end{center}
\end{table}


\subsection{Ablations}
\label{sec:ablations}


We conduct ablation studies in this section. We propose a tight fusion grounding model for open-set object detection and a sub-sentence level text prompt. To verify the effectiveness of the model design, we remove some fusion blocks for different variants. Results are shown in Table \ref{table:ablation}. All models are pre-trained on O365 with a Swin-T backbone. 

{The results show that encoder fusion significantly improves model performance on both COCO and LVIS datasets. The results from comparing model $\#$1 with the baseline model $\#$0 validate this observation. Other techniques, such as language-guided query selection, text cross-attention, and sub-sentence text prompt, also contribute positively to the LVIS performance, yielding significant gains of +3.0 AP, +1.8 AP, and +0.5 AP, respectively. Additionally, these methods enhance the COCO zero-shot performance, further underscoring their effectiveness. However, we observed that language-guided query selection and sub-sentence text prompt had minimal impact on the COCO fine-tune performance. This outcome is reasonable, given that these methods do not alter model parameters or add computational burdens. Text cross-attention, while introducing fewer parameters than encoder fusion, showed less performance improvement compared to encoder fusion (+0.6 vs. +0.8). This finding suggests that fine-tuning performance is predominantly influenced by the model's parameters, indicating that scaling models is a promising direction for enhancing performance.}


\begin{table}
\small
\resizebox{0.7\linewidth}{!}{%
\begin{tabu}{
l@{\hskip9pt} |  
l@{\hskip9pt} |  
c@{\hskip9pt} c@{\hskip9pt} |
c@{\hskip9pt}
}
 \toprule
 \multirow{2}{*}{\#ID}  & 
 \multirow{2}{*}{Model}  & 
 \multicolumn{2}{c|}{COCO minival} &
 \multicolumn{1}{c}{LVIS minival} 
 \\
  & 
  &
  \small{Zero-Shot} & \small{Fine-Tune} &
  \small{Zero-Shot}
  \\
\midrule
  0 & Grounding DINO (Full Model) & 46.7 & 56.9 & 16.1 \\
  \midrule[0.3pt]
  1 & w/o encoder fusion & 45.8 & 56.1  & 13.1 \\
  2 & static query selection & 46.3  & 56.6 &  13.6 \\
  3 & w/o text cross-attention & 46.1 & 56.3 & 14.3 \\
  4 & word-level text prompt & 46.4 & 56.6 & 15.6 \\
  \bottomrule
\end{tabu}}
\vspace{0.1cm}
\caption{Ablations for our model. All models are trained on the O365 dataset with a Swin Transformer Tiny backbone.}
\vspace{-0.3cm}
\label{table:ablation}
\end{table}

%


\section{Conclusion}
\label{sec:conclusion}


We have presented a Grounding DINO model in this paper. Grounding DINO extends DINO to open-set object detection, enabling it to detect arbitrary objects given texts as queries. We review open-set object detector designs and propose a tight fusion approach to better fusing cross-modality information. We propose a sub-sentence level representation to use detection data for text prompts in a more reasonable way. The results show the effectiveness of our model design and fusion approach. 
Moreover, we extend open-set object detection to REC tasks and perform evaluation accordingly. We show that existing open-set detectors do not work well for REC data without fine-tuning. Hence we call extra attention to REC zero-shot performance in future studies.

\noindent
\textbf{Limitations:} Despite the great performance on open-set object detection settings, Grounding DINO cannot be used for segmentation tasks like GLIPv2. Our training data is less than the largest GLIP model, which may limit our final performance. Moreover, we find that our model will produce false positive results in some cases, which may need more techniques or data to reduce the hallucination.

\noindent
\textbf{Social Impacts:} The use of deep learning models, such as this one, exposes them to vulnerabilities through adversarial attacks. Additionally, the accuracy and correctness of the model's outputs cannot be guaranteed. There is also the risk that the open-set detection capabilities of the model could be exploited for unlawful purposes.



\section*{Acknowledgement}
We thank the authors of GLIP \cite{li2021grounded}: Liunian Harold Li, Pengchuan Zhang, and Haotian Zhang for their helpful discussions and instructions. We also thank Tiancheng Zhao, the author of OmDet \cite{TianchengZhao2022OmDetLO}, and Jianhua Han, the author of DetCLIP \cite{LeweiYao2022DetCLIPDV}, for their response to their model details. We thank He Cao of The Hong Kong University of Science and Technology for his help on diffusion models.

%
%
\bibliographystyle{splncs04}
\bibliography{main}

\clearpage
\appendix

\section{More Implementation Details}
\label{sec:imple_details}


\subsection{Hyperparameters}
Table \ref{tab:hyperparameters} presents the hyperparameters used in our main experiments.

\begin{table}[ht]
    \centering
    \begin{tabular}{l|c}
	\shline
	Item & Value \\
	\shline
        optimizer   & AdamW \\ 
	lr & 1e-4 \\ 
	lr of image backbone & 1e-5 \\ 
	lr of text backbone & 1e-5 \\ 
	weight decay & 0.0001 \\ 
	clip max norm & 0.1 \\ 
	number of encoder layers & 6 \\ 
	number of decoder layers & 6 \\ 
	dim feedforward & 2048 \\ 
	hidden dim & 256 \\ 
	dropout & 0.0 \\ 
	nheads & 8 \\ 
	number of queries & 900 \\ 
	set cost class & 1.0 \\ 
	set cost bbox & 5.0 \\ 
	set cost giou & 2.0 \\ 
	ce loss coef & 2.0 \\ 
	bbox loss coef & 5.0 \\ 
	giou loss coef & 2.0 \\ 
	\shline
    \end{tabular}
    \vspace{0.2cm}
    \caption{Hyper-parameters used in our pre-trained models.}
    \label{tab:hyperparameters}
\end{table}

\subsection{Pseudo Code Language-Guided Query Selection}
We present the pseudo-code of the Language-Guided Query Selection module in Algorithm \ref{code:qs}.




\begin{algorithm}[ht]
\scriptsize
\caption{Pseudocode of Language-guided Query Selection in PyTorch-like style. }
\label{code:qs}
\definecolor{codeblue}{rgb}{0.25,0.5,0.5}
\definecolor{codegreen}{rgb}{0,0.6,0}
\definecolor{codekw}{rgb}{0.85, 0.18, 0.50}
\lstset{
  backgroundcolor=\color{white},
  basicstyle=\fontsize{7.5pt}{7.5pt}\ttfamily\selectfont,
  columns=fullflexible,
  breaklines=true,
  captionpos=b,
  commentstyle=\fontsize{7.5pt}{7.5pt}\color{codegreen},
  keywordstyle=\fontsize{7.5pt}{7.5pt}\color{codekw},
  escapechar={|}, 
}
\begin{lstlisting}[language=Python]
"""
Input:
image_feat: (bs, num_img_tokens, ndim)
text_feat: (bs, num_text_tokens, ndim)
num_query: int

Output:
topk_idx: (bs, num_query)
"""

logits = torch.einsum("bic,btc->bit", image_feat, text_feat) # bs, num_img_tokens, num_text_tokens
logits_per_img_feat = logits.max(-1)[0]# bs, num_img_tokens
topk_idx = torch.topk(logits_per_img_feature, num_query, dim=1)[1] # bs, num_query
\end{lstlisting}
\end{algorithm}

The variables \texttt{image\_feat} and \texttt{text\_feat} are used for image and text features, respectively. \texttt{num\_query} is the number of queries in the decoder, which is set to $900$ in our implementation. We use \texttt{bs} and \texttt{ndim} for batch size and feature dimension in the pseudo-code. \texttt{num\_img\_tokens} and \texttt{num\_text\_tokens} are used for the number of image and text tokens, respectively. 

\section{Data Usage}
\label{sec:data_usage}
We use three types of data in our model pre-train. 
\begin{enumerate}
    \item \textbf{Detection data.} Following GLIP \cite{li2021grounded}, we reformulate the object detection task to a phrase grounding task by concatenating the category names into text prompts. We use COCO \cite{lin2014microsoft}, O365 \cite{shao2019objects365}, and OpenImage(OI) \cite{krasin2017openimages} for our model pretrain. To simulate different text inputs, we randomly sampled category names from all categories in a dataset on the fly during training.
    \item \textbf{Grounding data.} We use the GoldG and RefC data as grounding data. Both GoldG and RefC are preprocessed by MDETR \cite{kamath2021mdetr}. These data can be fed into Grounding DINO directly. GoldG contains images in  Flickr30k entities \cite{plummer2015flickr30k, BryanAPlummer2015Flickr30kEC} and Visual Genome \cite{RanjayKrishna2017VisualGC}. RefC contains images in RefCOCO, RefCOCO+, and RefCOCOg. 
    \item \textbf{Caption data.} To enhance the model performance on novel categories, we feed the semantic-rich caption data to our model. Following GLIP, we use the pseudo-labeled caption data for model training. In our experiments, we use the same data with GLIP under comparable settings. More specifically, we use GLIP-T annotated caption data for Grounding DINO T, while GLIP-L annotated caption data for Grounding DINO L.
\end{enumerate}

There are two versions of the O365 dataset, which we termed O365v1 and O365v2, respectively. O365v1 is a subset of O365v2. O365v1 contains about 600K images, while O365v2 contains about 1.7M images. Following previous works \cite{li2021grounded, LeweiYao2022DetCLIPDV}, we pre-train the Grounding DINO T on O365v1 for a fair comparison. The Grounding DINO L is pre-trained on O365v2 for a better result.

\section{More Experiment Results}

\subsection{Transfer from DINO to Grounding DINO}
\label{sec:dino_to_groundingdino}

Recent work has presented many large-scale image models for detection with DINO architecture\footnote{See model instances at \url{https://github.com/IDEA-Research/detrex} }. It is computationally expensive to train a Grounding DINO model from scratch. However, the cost can be significantly reduced if we leverage pre-trained DINO weights. Hence, we conduct some experiments to transfer pre-trained DINO to Grounding DINO models. We freeze the modules co-existing in DINO and Grounding DINO and fine-tune the other parameters only. (We compare DINO and Grounding DINO in Sec. \ref{sec:comparison_dino_groundingdino}.) The results are available in Table \ref{table:dino2grounding}. 

\begin{table}
\small
\resizebox{0.8\linewidth}{!}{
\begin{tabu}{
l@{\hskip9pt} |  
c@{\hskip9pt} c@{\hskip9pt}  |  
c@{\hskip9pt} |
c@{\hskip9pt} | 
c@{\hskip9pt} 
}
 \toprule
 \multirow{2}{*}{Model}  & 
 \multicolumn{2}{c|}{Pre-Train Data} &
 \multicolumn{1}{c|}{COCO minival} &
 \multicolumn{1}{c|}{LVIS minival} &
 \multicolumn{1}{c}{ODinW} 
 \\
  &
  \small{DINO Pre-Train} & \small{Grounded Fine-Tune} &
  Zero-Shot &
  Zero-Shot &
  Zero-Shot
  \\
  \midrule
  Grounding DINO T  & - & O365 & 46.7 & 16.2 & 14.5 \\
  (from scratch)  & - & O365,GoldG & 48.1 & 25.6 & 20.0  \\
  \midrule
  Grounding DINO T  & O365 & O365          & 46.5 & 17.9 & 13.6 \\
  (from pre-trained DINO)  & O365 & O365,GoldG    & 46.4 & 26.1 & 18.5 \\
  \bottomrule
\end{tabu}}
\caption{Transfer pre-trained DINO to Grounding DINO. We freeze shared modules between DINO and Grounding DINO during grounded fine-tuning. All models are trained with a Swin Transformer Tiny backbone. }
\label{table:dino2grounding}
\end{table}

It shows that we can achieve similar performances with Grounding DINO-Training only text and fusion blocks using a pre-trained DINO. Interestingly, the DINO-pre-trained Grounding DINO outperforms the standard Grounding DINO on LVIS under the same setting. 
The results show that there might be much room for model training improvement, which will be our future work to explore.

\begin{table}
    \centering
    \includegraphics[width=0.45\textwidth]{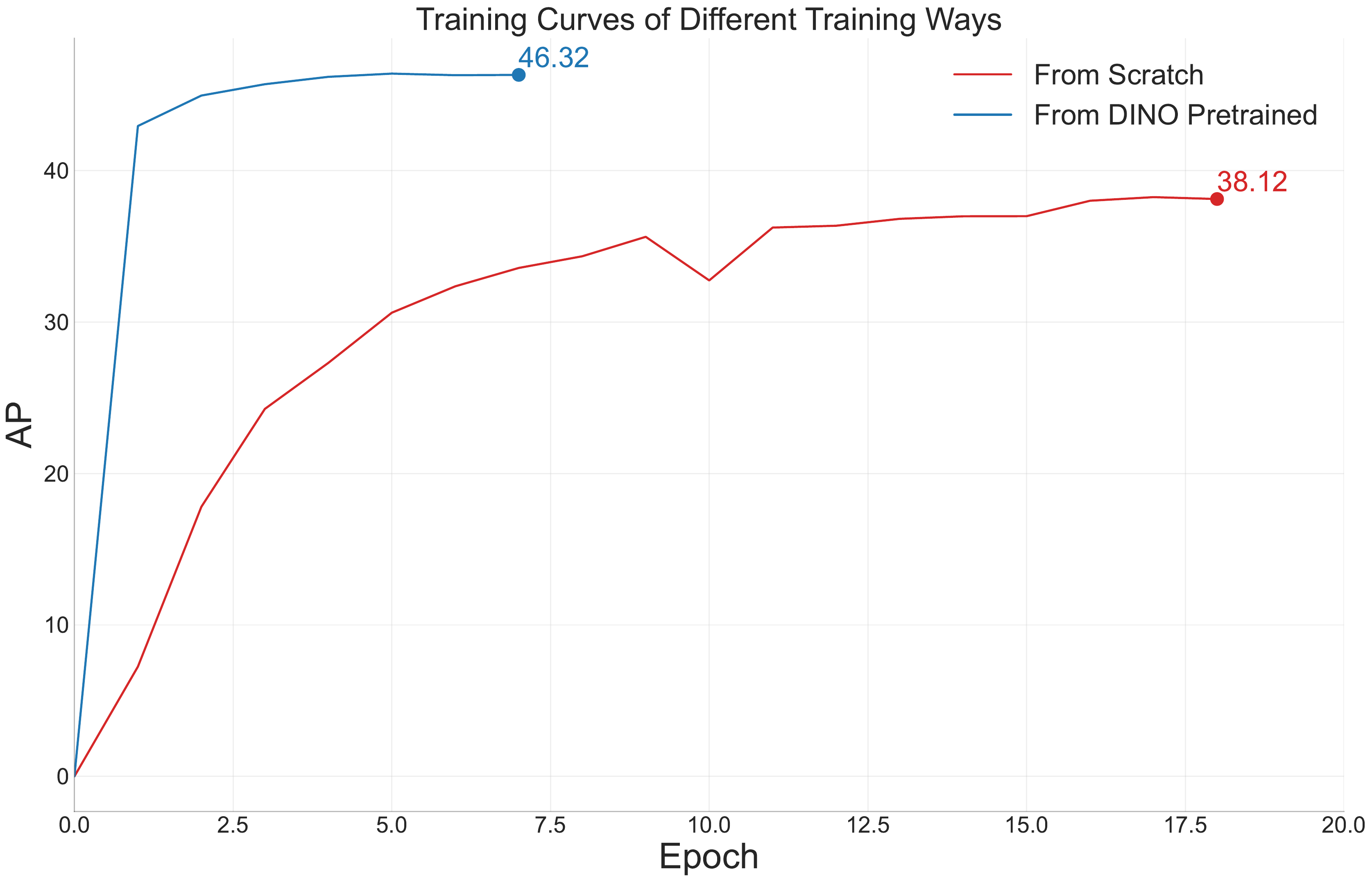}
    \caption{
    Comparison between two Grounding DINO variants: Training from scratch and transfer from DINO-pretrained models. The models are trained on O365 and evaluated on COCO.
    }
    \label{fig:curve_fewshot}
\end{table}

With a pre-trained DINO initialization, the model converges faster than Grounding DINO from scratch, as shown in Fig. \ref{fig:curve_fewshot}. 
Notably, we use the results without exponential moving average (EMA) for the curves in Fig. \ref{fig:curve_fewshot}, which results in a different final performance that in Table \ref{table:dino2grounding}. As the model trained from scratch need more training time, we only show results of early epochs.

\begin{figure*}[t]
    \centering
    \includegraphics[width=\linewidth]{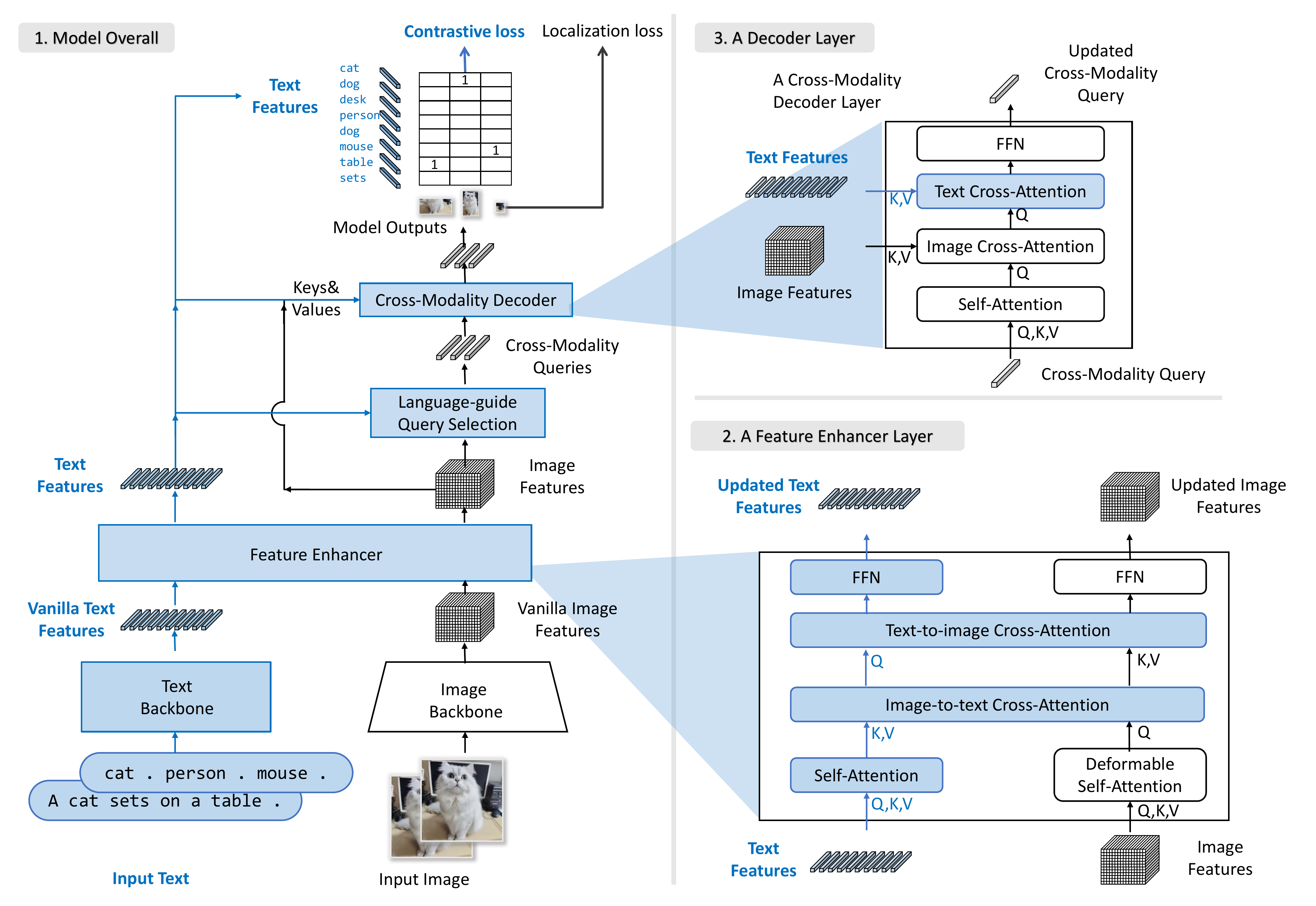}
    \caption{Comparison between DINO and our Grounding DINO. We mark the modifications in blue. Best view in color.}
    \label{fig:dino_and_groundingdino}
\end{figure*}

\textbf{Comparison between DINO and Grounding DINO}
\label{sec:comparison_dino_groundingdino}
To illustrate the difference between DINO and Grounding DINO, we compare DINO and Grounding DINO in Fig. \ref{fig:dino_and_groundingdino}. We mark the DINO blocks in gray, while the newly proposed modules are shaded in blue.

\subsection{Detailed Results on COCO Detection Benchmarks}

\textbf{COCO Detection Results under the 1$\times$ Setting}
\label{sec:coco}

\begin{table*}[h]
    \centering
    \vspace{-0.3cm}
    \resizebox{0.82\textwidth}{!}{%
    \begin{tabular}{l|c|cccccc}
        \toprule
        Model  & Epochs & AP & AP$_{50}$ & AP$_{75}$ & AP$_{S}$ & AP$_{M}$ & AP$_{L}$  \\
        \midrule
        Faster-RCNN(5scale) \cite{ren2015faster}& $12$ & $37.9$ & $58.8$ & $41.1$ & $22.4$ & $41.1$ & $49.1$ \\
        DETR(DC5) \cite{carion2020end}& $12$ & $15.5$ & $29.4$ & $14.5$ & $4.3$ & $15.1$ & $26.7$  \\
        Deformable DETR(4scale)\cite{zhu2020deformable}& $12$ & $41.1$ &  $-$ & $-$ & $-$ & $-$ & \\
        DAB-DETR(DC5)$^\dag$ \cite{liu2022dabdetr}& $12$ & $38.0$ & $60.3$ & $39.8$ & $19.2$ & $40.9$ & $55.4$  \\
        Dynamic DETR(5scale) \cite{Dai_2021_ICCV}& $12$ & $42.9$ & $61.0$ & $46.3$ & $24.6$ & $44.9$ & $54.4$ \\
        Dynamic Head(5scale) \cite{dai2021dynamic}& $12$ & $43.0$ &$60.7$ & $46.8$ & $24.7$ & $46.4$ & $53.9$ \\
        HTC(5scale) \cite{chen2019hybrid}& $12$ & $42.3$ & $-$ & $-$ & $-$ & $-$ & $-$ \\
        DN-Deformable-DETR(4scale) \cite{li2022dn}& $12$ & $43.4$ & $61.9$ & $47.2$ & $24.8$ & $46.8$ & $59.4$ \\
        DINO-4scale \cite{zhang2022dino}    &$12$ & ${49.0}$& ${66.6}$ & ${53.5}$ & ${32.0}$ & ${52.3}$ & ${63.0}$ \\
        \midrule
        Grounding DINO (4scale)& $12$ & $48.1$ & $65.8$ & $52.3$ & $30.4$ & $51.3$ & $62.3$ \\
        \bottomrule
    \end{tabular}}
    \centering
    \caption{{Results for Grounding DINO and other detection models with the ResNet50 backbone on COCO \texttt{val2017} trained with $12$ epochs (the so called $1\times$ setting). 
    }}
    \label{tab:12ep}
    \vspace{-.1cm}
\end{table*}

We present the performance of Grounding DINO on standard COCO detection benchmark in Table \ref{tab:12ep}. All models are trained with a ResNet-50 \cite{he2015deep} backbone for 12 epochs. 
Grounding DINO achieves 48.1 AP under the research setting, which shows that Grounding DINO is a strong closed-set detector.
However, it is inferior compared with the original DINO. We suspect that the new components may make the model harder to optimize than DINO.

\subsection{Detailed Results on ODinW Benchmarks}
We present detailed results of Grounding DINO on ODinW35\cite{ChunyuanLi2022ELEVATERAB} in Table \ref{table:odinw_detail_swint_og}, Table \ref{table:odinw_detail_swint_ogc}, and Table \ref{table:odinw_detail_swinl}.

\begin{table}[ht]
\resizebox{0.7\columnwidth}{!}{%
\begin{tabular}{l|cccccc}
\toprule
Dataset  & AP & AP$_{50}$ & AP$_{75}$ & AP$_{S}$ & AP$_{M}$ & AP$_{L}$  \\
\midrule
AerialMaritimeDrone\_large   & 9.48  & 15.61 & 8.35  & 8.72  & 10.28 & 2.91  \\
AerialMaritimeDrone\_tiled   & 17.56 & 26.35 & 13.89 & 0     & 1.61  & 28.7  \\
AmericanSignLanguageLetters  & 1.45  & 2.21  & 1.39  & -1    & -1    & 1.81  \\
Aquarium                     & 18.83 & 34.32 & 18.19 & 10.65 & 20.64 & 21.52 \\
BCCD\_BCCD                   & 6.17  & 11.31 & 6.04  & 1.27  & 9.09  & 6.89  \\
ChessPiece                   & 6.99  & 11.13 & 9.03  & -1    & -1    & 8.11  \\
CottontailRabbits            & 71.93 & 85.05 & 85.05 & -1    & 70    & 73.58 \\
DroneControl\_Drone\_Control & 6.15  & 10.95 & 6.23  & 2.08  & 6.91  & 6.16  \\
EgoHands\_generic            & 48.07 & 75.06 & 56.52 & 1.48  & 11.42 & 51.84 \\
EgoHands\_specific           & 0.66  & 1.25  & 0.64  & 0     & 0.02  & 0.92  \\
HardHatWorkers               & 2.39  & 9.17  & 1.07  & 2.13  & 4.32  & 4.6   \\
MaskWearing                  & 0.58  & 1.43  & 0.56  & 0.12  & 0.51  & 4.66  \\
MountainDewCommercial        & 18.22 & 29.73 & 21.33 & 0     & 23.23 & 49.8  \\
NorthAmericaMushrooms        & 65.48 & 71.26 & 66.18 & -1    & -1    & 65.49 \\
OxfordPets\_by-breed         & 0.27  & 0.6   & 0.21  & -1    & 1.38  & 0.33  \\
OxfordPets\_by-species       & 1.66  & 5.02  & 1     & -1    & 0.65  & 1.89  \\
PKLot\_640                   & 0.08  & 0.26  & 0.02  & 0.14  & 0.79  & 0.11  \\
Packages                     & 56.34 & 68.65 & 68.65 & -1    & -1    & 56.34 \\
PascalVOC                    & 47.21 & 57.59 & 51.28 & 16.53 & 39.51 & 58.5  \\
Raccoon\_Raccoon             & 44.82 & 76.44 & 46.16 & -1    & 17.08 & 48.56 \\
ShellfishOpenImages          & 23.08 & 32.21 & 26.94 & -1    & 18.82 & 23.28 \\
ThermalCheetah               & 12.9  & 19.65 & 14.72 & 0     & 8.35  & 50.15 \\
UnoCards                     & 0.87  & 1.52  & 0.96  & 2.91  & 2.18  & -1    \\
VehiclesOpenImages           & 59.24 & 71.88 & 64.69 & 7.42  & 32.38 & 72.21 \\
WildfireSmoke                & 25.6  & 43.96 & 25.34 & 5.03  & 18.85 & 42.59 \\
boggleBoards                 & 0.81  & 2.92  & 0.12  & 2.96  & 1.13  & -1    \\
brackishUnderwater           & 1.3   & 1.88  & 1.4   & 0.99  & 1.75  & 11.39 \\
dice\_mediumColor            & 0.16  & 0.72  & 0.07  & 0.38  & 3.3   & 2.23  \\
openPoetryVision             & 0.18  & 0.5   & 0.06  & -1    & 0.25  & 0.17  \\
pistols                      & 46.4  & 66.47 & 47.98 & 4.51  & 22.94 & 55.03 \\
plantdoc                     & 0.34  & 0.51  & 0.35  & -1    & 0.28  & 0.86  \\
pothole                      & 19.87 & 28.94 & 22.23 & 12.49 & 15.6  & 28.78 \\
selfdrivingCa                & 9.46  & 19.13 & 8.19  & 0.85  & 6.82  & 16.51 \\
thermalDogsAndPeople         & 72.67 & 86.65 & 79.98 & 33.93 & 30.2  & 86.71 \\
websiteScreenshots           & 1.51  & 2.8   & 1.42  & 0.85  & 2.06  & 2.59 \\
\bottomrule
\end{tabular}}
\caption{Detailed results on 35 datasets in ODinW  of Grounding DINO with Swin-T pre-trained on O365 and GoldG.}
\label{table:odinw_detail_swint_og}
\end{table}
\begin{table}[]
\resizebox{0.7\columnwidth}{!}{%
\begin{tabular}{l|cccccc}
\toprule
Dataset  & AP & AP$_{50}$ & AP$_{75}$ & AP$_{S}$ & AP$_{M}$ & AP$_{L}$  \\
\midrule
AerialMaritimeDrone\_large   & 10.3  & 18.17 & 9.21  & 8.92  & 11.2  & 7.35  \\
AerialMaritimeDrone\_tiled   & 17.5  & 28.04 & 18.58 & 0     & 3.64  & 24.16 \\
AmericanSignLanguageLetters  & 0.78  & 1.17  & 0.76  & -1    & -1    & 1.02  \\
Aquarium                     & 18.64 & 35.27 & 17.29 & 11.33 & 17.8  & 21.34 \\
BCCD\_BCCD                   & 11.96 & 22.77 & 8.65  & 0.16  & 5.02  & 13.15 \\
ChessPiece                   & 15.62 & 22.02 & 20.19 & -1    & -1    & 15.72 \\
CottontailRabbits            & 67.61 & 78.82 & 78.82 & -1    & 70    & 68.09 \\
DroneControl\_Drone\_Control & 4.99  & 8.76  & 5     & 0.65  & 5.03  & 8.61  \\
EgoHands\_generic            & 57.64 & 90.18 & 66.78 & 3.74  & 24.67 & 61.33 \\
EgoHands\_specific           & 0.69  & 1.37  & 0.63  & 0     & 0.02  & 1.03  \\
HardHatWorkers               & 4.05  & 13.16 & 1.96  & 2.29  & 7.55  & 9.81  \\
MaskWearing                  & 0.25  & 0.81  & 0.15  & 0.09  & 0.13  & 2.78  \\
MountainDewCommercial        & 25.46 & 39.08 & 28.89 & 0     & 32.53 & 58.38 \\
NorthAmericaMushrooms        & 68.18 & 72.89 & 69.75 & -1    & -1    & 68.62 \\
OxfordPets\_by-breed         & 0.21  & 0.42  & 0.22  & -1    & 2.91  & 0.17  \\
OxfordPets\_by-species       & 1.3   & 3.95  & 0.71  & -1    & 0.28  & 1.62  \\
PKLot\_640                   & 0.06  & 0.18  & 0.02  & 0.03  & 0.59  & 0.15  \\
Packages                     & 60.53 & 76.24 & 76.24 & -1    & -1    & 60.53 \\
PascalVOC                    & 55.65 & 66.51 & 60.47 & 19.61 & 44.25 & 67.21 \\
Raccoon\_Raccoon             & 60.07 & 84.81 & 66.5  & -1    & 11.23 & 65.86 \\
ShellfishOpenImages          & 29.56 & 38.08 & 33.5  & -1    & 6.38  & 29.95 \\
ThermalCheetah               & 17.72 & 25.93 & 19.61 & 1.04  & 20.02 & 63.69 \\
UnoCards                     & 0.81  & 1.3   & 1     & 2.6   & 1.01  & -1    \\
VehiclesOpenImages           & 58.49 & 71.56 & 63.64 & 8.22  & 28.03 & 71.1  \\
WildfireSmoke                & 20.04 & 39.74 & 22.49 & 4.13  & 15.71 & 30.41 \\
boggleBoards                 & 0.29  & 1.15  & 0.04  & 1.8   & 0.57  & -1    \\
brackishUnderwater           & 1.47  & 2.34  & 1.58  & 2.32  & 3.31  & 9.96  \\
dice\_mediumColor            & 0.33  & 1.38  & 0.15  & 0.03  & 1.05  & 12.57 \\
openPoetryVision             & 0.05  & 0.19  & 0     & -1    & 0.09  & 0.21  \\
pistols                      & 66.99 & 86.34 & 72.65 & 16.25 & 39.24 & 75.98 \\
plantdoc                     & 0.36  & 0.47  & 0.39  & -1    & 0.24  & 0.82  \\
pothole                      & 25.21 & 38.21 & 26.01 & 8.94  & 18.45 & 39.28 \\
selfdrivingCa                & 9.95  & 20.55 & 8.28  & 1.36  & 7.27  & 15.46 \\
thermalDogsAndPeople         & 67.89 & 80.85 & 78.66 & 45.05 & 30.24 & 85.56 \\
websiteScreenshots           & 1.3   & 2.26  & 1.21  & 0.95  & 1.81  & 2.23 \\
\bottomrule
\end{tabular}}
\caption{Detailed results on 35 datasets in ODinW of Grounding DINO with Swin-T pre-trained on O365, GoldG, and Cap4M.}
\label{table:odinw_detail_swint_ogc}
\end{table}
\begin{table}[]
\resizebox{0.7\columnwidth}{!}{%
\begin{tabular}{l|cccccc}
\toprule
Dataset  & AP & AP$_{50}$ & AP$_{75}$ & AP$_{S}$ & AP$_{M}$ & AP$_{L}$  \\
\midrule
AerialMaritimeDrone\_large   & 12.64 & 18.44 & 14.75 & 9.15  & 19.16 & 0.98  \\
AerialMaritimeDrone\_tiled   & 20.47 & 34.81 & 12.79 & 0     & 7.61  & 26.93 \\
AmericanSignLanguageLetters  & 3.94  & 4.84  & 4     & -1    & -1    & 4.48  \\
Aquarium                     & 28.14 & 45.47 & 30.97 & 12.1  & 24.71 & 39.42 \\
BCCD\_BCCD                   & 23.85 & 36.92 & 28.88 & 0.3   & 10.8  & 24.43 \\
ChessPiece                   & 18.44 & 26.3  & 23.33 & -1    & -1    & 18.62 \\
CottontailRabbits            & 71.66 & 88.48 & 88.48 & -1    & 66    & 73.04 \\
DroneControl\_Drone\_Control & 7.16  & 11.56 & 7.67  & 2.29  & 10.6  & 7.68  \\
EgoHands\_generic            & 52.08 & 81.57 & 59.15 & 1.12  & 31.78 & 55.46 \\
EgoHands\_specific           & 1.22  & 2.28  & 1.2   & 0     & 0.05  & 1.5   \\
HardHatWorkers               & 9.14  & 23.64 & 5.6   & 5.09  & 15.34 & 13.59 \\
MaskWearing                  & 1.64  & 4.69  & 1.18  & 0.44  & 1.05  & 8.67  \\
MountainDewCommercial        & 33.28 & 53.59 & 32.76 & 0     & 35.86 & 80    \\
NorthAmericaMushrooms        & 72.33 & 73.18 & 73.18 & -1    & -1    & 72.39 \\
OxfordPets\_by-breed         & 0.58  & 1.05  & 0.59  & -1    & 4.46  & 0.6   \\
OxfordPets\_by-species       & 1.64  & 4.8   & 0.87  & -1    & 1.51  & 1.8   \\
PKLot\_640                   & 0.25  & 0.71  & 0.05  & 0.31  & 1.44  & 0.4   \\
Packages                     & 63.86 & 76.24 & 76.24 & -1    & -1    & 63.86 \\
PascalVOC                    & 66.01 & 76.65 & 71.8  & 32.01 & 55.7  & 75.37 \\
Raccoon\_Raccoon             & 65.81 & 90.39 & 69.93 & -1    & 26    & 68.97 \\
ShellfishOpenImages          & 62.47 & 74.25 & 70.07 & -1    & 26    & 63.06 \\
ThermalCheetah               & 21.33 & 26.11 & 24.92 & 2.39  & 15.84 & 75.34 \\
UnoCards                     & 0.52  & 0.84  & 0.66  & 3.02  & 0.92  & -1    \\
VehiclesOpenImages           & 62.74 & 75.15 & 67.23 & 10.66 & 47.46 & 76.36 \\
WildfireSmoke                & 23.66 & 45.72 & 25.06 & 1.58  & 22.22 & 35.27 \\
boggleBoards                 & 0.28  & 1.04  & 0.05  & 5.64  & 0.7   & -1    \\
brackishUnderwater           & 2.41  & 3.39  & 2.79  & 4.43  & 3.88  & 21.22 \\
dice\_mediumColor            & 0.26  & 1.15  & 0.03  & 0     & 1.09  & 4.07  \\
openPoetryVision             & 0.08  & 0.35  & 0.01  & -1    & 0.15  & 0.11  \\
pistols                      & 71.4  & 90.69 & 77.21 & 18.74 & 39.58 & 80.78 \\
plantdoc                     & 2.02  & 2.64  & 2.37  & -1    & 0.5   & 2.82  \\
pothole                      & 30.4  & 44.22 & 33.84 & 12.27 & 18.84 & 48.57 \\
selfdrivingCa                & 9.25  & 17.72 & 8.39  & 1.93  & 7.03  & 13.02 \\
thermalDogsAndPeople         & 72.02 & 86.02 & 79.47 & 29.16 & 68.05 & 86.75 \\
websiteScreenshots           & 1.32  & 2.64  & 1.16  & 0.79  & 1.8   & 2.46  \\
\bottomrule
\end{tabular}}
\caption{Detailed results on 35 datasets in ODinW of Grounding DINO with Swin-L pre-trained on O365, OI, GoldG, Cap4M, COCO, and RefC.}
\label{table:odinw_detail_swinl}
\end{table}

\subsection{Model Efficiency}
\label{sec:efficiency}
We compare the model size and efficiency between Grounding DINO T and GLIP-T in Table \ref{table:gflops}. The results show that our model has a smaller parameter size and better efficiency than GLIP. 

\subsection{{Ablations for More Decoder Queries}}
\label{sec:query_number}

{
To verify the model performance with more decoder queries, we conducted additional experiments with 1200 and 1500 queries on the COCO and LVIS datasets, detailed in Table \ref{tab:morequery} below. 



The results indicate that models with 1200 and 1500 queries slightly outperform the 900-query version on LVIS rare classes, suggesting better coverage of rare classes. However, the improvement is marginal, as the 900-query model already sufficiently covers all objects in both COCO and LVIS. Additionally, introducing more queries exacerbates data imbalance during training, as the model is trained on objects from sampled categories. This imbalance could offset the benefits of additional queries.
}

\begin{table*}[h]
    \centering
    \resizebox{0.8\textwidth}{!}{%
    \begin{tabular}{c|c|c|c|c|c|c|c|c|c}
    \toprule
    \multirow{2}{*}{Pretrain Data} & \multirow{2}{*}{Query Num} & \multicolumn{4}{c}{\textbf{COCO}} & \multicolumn{4}{c}{\textbf{LVIS}} 
    \\
    \cline{3-10}
    &  & {AP} & {AP$_S$} & {AP$_M$} & {AP$_L$} & {AP} & {AP$_r$} & {AP$_c$} & {AP$_f$} \\ 
    \midrule
    O365 & 900 & 46.7 & 32.1 & 50.0 & 61.3 & 15.8 & 9.4 & 22.9 & 28.4 \\ 
    O365 & 1200 & 46.7 & 32.3 & 49.9 & 61.1 & 15.7 & 9.0 & 23.1 & 29.1 \\
    O365 & 1500 & 46.9 & 32.7 & 50.3 & 61.3 & 15.8 & 9.2 & 23.0 & 29.0 \\ 
    \bottomrule
    \end{tabular}}
    \centering
    \caption{{Results for Grounding DINO Tiny with more decoder queries
    }}
    \label{tab:morequery}
    \vspace{-.1cm}
\end{table*}

\subsection{{Results with Different Language Encoder for REC}}

{
To verify the impacts of language encoders with different sizes, we conducted experiments using two variants of BERT: bert-base-uncased (BERT-B) and bert-large-uncased (BERT-L). These models were trained on a combined dataset consisting of RefCOCO, RefCOCO+, and RefCOCOg. We removed the leaked data in the combined dataset for a fair comparison. It's important to note that training on this combined dataset, as opposed to tuning each dataset separately, might result in slightly lower performance. For a fair comparison, we initialized the models with the O365+GoldG+Cap4M checkpoint, except for the BERT parameters. Limited by time and resources, the training duration was capped at 18 epochs.

Interestingly, our results showed that Grounding DINO with BERT-B outperformed or matched the BERT-L variant in most metrics (as shown in Table \ref{tab:bertl}). This suggests that our default use of BERT-B during the pretrain stage may have contributed to its better performance. Moreover, we didn't observe significant improvements in the late stages of training, indicating that both models were nearing their optimal performance.

This outcome suggests that the main limitation in enhancing REC performance lies within the detection branch, rather than the language processing module. A dedicated model for REC data might be helpful for REC tasks.}

\begin{table*}[h]
    \centering
    \resizebox{0.8\textwidth}{!}{%
    \begin{tabular}{l|ccc|ccc|cc}
    \toprule
    \multirow{2}{*}{Model} & \multicolumn{3}{c|}{RefCOCO} & \multicolumn{3}{c|}{RefCOCO+} & \multicolumn{2}{c}{RefCOCOg} \\
    \cline{2-9}
    & val & testA & testB & val & testA & testB & val & test \\
    \midrule 
    Grounding DINO T (BERT-B) & 87.4 & 91.6 & 84.2 & 78.6 & 86.5 & 73.4 & 81.6 & 83.3   \\
    Grounding DINO T (BERT-L) & 87.0 & 91.4 & 84.0 & 78.1 & 86.4 & 73.0 & 81.7 & 83.3 \\
    \bottomrule
    \end{tabular}}
    \centering
    \caption{{Results for Grounding DINO with different language encoders 
    }}
    \label{tab:bertl}
    \vspace{-.1cm}
\end{table*}

\subsection{{Detic Pseudo-labeled Data for LVIS}}

To illustrate the potential of our model under similar conditions, we conducted oracle experiments as shown in Table \ref{tab:detic}. We pseudo-labeled the ImageNet dataset using a pre-trained Detic\cite{zhou2022detecting} model and filtered out LVIS-related images for training, creating a new pseudo-labeled dataset named IN22K-LVIS-1M. It contains about 1M pseudo-labeled images for training. Note that our model under the setting may be not a real zero-shot setting, as our pseudo labeler Detic is trained with LVIS data. The results from these experiments suggest that Grounding DINO can achieve promising LVIS performance, even without direct LVIS training data. A similarly distributed dataset can help the model to generalize well.

\begin{table*}[h]
    \centering
    \resizebox{\textwidth}{!}{%
    \begin{tabular}{c|c|cc}
    \toprule
    \multirow{2}{*}{Model} & \multirow{2}{*}{Pre-train Data} & \multicolumn{2}{c}{LVIS MiniVal - AP(r/c/f)}  \\
     &  & Zero-shot & Fine-tune\\
    \midrule
    DetCLIPv2-T  & OG + CC15M & 40.4 (36.0 / 41.7 / 40.0) & 50.7 (44.3 / 52.4 / 50.3) \\
    Grounding DINO T  & OG+IN22K-LVIS-1M & 40.6 (38.5 / 41.1 / 40.4) & 54.5 (47.3 / 53.9 / 56.1)\\
    \hline
    \end{tabular}}
    \centering
    \caption{{Oracle experiments on LVIS. Note that our model under the setting may be not a real zero-shot setting, as our pseudo labeler Detic is trained with LVIS data. }}
    \label{tab:detic}
    \vspace{-.1cm}
\end{table*}

\subsection{{Comparison between Grounding DINO and GLIP on ODinW}}
{
In our comparison of Grounding DINO and GLIP across various datasets in ODinW, as presented in Table \ref{tab:comp_odinw}, we observe that Grounding DINO underperforms on certain uncommon datasets where both models generally show limited effectiveness. For instance, in the PlantDoc dataset, Grounding DINO scores 0.36 compared to GLIP's 1.1. This dataset includes infrequent categories such as "Tomato leaf mosaic virus," which are not well-represented in the training data. These findings highlight the need for improving data quality to enhance overall model performance.
}

\begin{table*}[!htbp]
    \centering
    \resizebox{\textwidth}{!}{%
    \begin{tabular}{l
    >{\columncolor[HTML]{FFFFFF}}l l}
    \toprule
    Metric                                                   & GLIP-T & Grounding DINO T              \\ 
    \midrule
    Average Score (↑)                                                  & 19.6   & \cellcolor[HTML]{FFFFFF}22.3  \\
    Median Score (↑)                                                   & 5.1    & \cellcolor[HTML]{FFFFFF}11.9  \\
    \midrule
    AerialMaritimeDrone\_large (↑)                                     & 13.70  & \cellcolor[HTML]{FFFFFF}10.30 \\
    AerialMaritimeDrone\_tiled (↑)                                     & 12.60  & \cellcolor[HTML]{FFFFFF}17.50 \\
    AmericanSignLanguageLetters\_American\_Sign\_Language\_Letters (↑) & 2.50   & \cellcolor[HTML]{FFFFFF}0.78  \\
    Aquarium\_Aquarium\_Combined (↑)                                   & 18.30  & \cellcolor[HTML]{FFFFFF}18.64 \\
    BCCD\_BCCD (↑)                                                     & 1.00   & \cellcolor[HTML]{FFFFFF}11.96 \\
    ChessPieces\_Chess\_Pieces (↑)                                     & 10.00  & \cellcolor[HTML]{FFFFFF}15.62 \\
    CottontailRabbits (↑)                                              & 69.70  & \cellcolor[HTML]{FFFFFF}67.61 \\
    DroneControl\_Drone\_Control (↑)                                   & 5.10   & \cellcolor[HTML]{FFFFFF}4.99  \\
    EgoHands\_generic (↑)                                              & 50.00  & \cellcolor[HTML]{FFFFFF}57.64 \\
    EgoHands\_specific (↑)                                             & 0.80   & \cellcolor[HTML]{FFFFFF}0.69  \\
    HardHatWorkers (↑)                                                 & 3.00   & \cellcolor[HTML]{FFFFFF}4.05  \\
    MaskWearing (↑)                                                    & 1.10   & \cellcolor[HTML]{FFFFFF}0.25  \\
    MountainDewCommercial (↑)                                          & 21.60  & \cellcolor[HTML]{FFFFFF}25.46 \\
    NorthAmericaMushrooms\_North\_American\_Mushrooms (↑)              & 75.10  & \cellcolor[HTML]{FFFFFF}68.18 \\
    OxfordPets\_by-breed (↑)                                           & 0.40   & \cellcolor[HTML]{FFFFFF}0.21  \\
    OxfordPets\_by-species (↑)                                         & 1.10   & \cellcolor[HTML]{FFFFFF}1.30  \\
    PKLot\_640 (↑)                                                     & 0.00   & \cellcolor[HTML]{FFFFFF}0.06  \\
    Packages\_Raw (↑)                                                  & 72.30  & \cellcolor[HTML]{FFFFFF}60.53 \\
    PascalVOC (↑)                                                      & 56.10  & \cellcolor[HTML]{FFFFFF}55.65 \\
    Raccoon\_Raccoon (↑)                                               & 57.80  & \cellcolor[HTML]{FFFFFF}60.07 \\
    ShellfishOpenImages (↑)                                            & 25.90  & \cellcolor[HTML]{FFFFFF}29.56 \\
    ThermalCheetah (↑)                                                 & 2.70   & \cellcolor[HTML]{FFFFFF}17.72 \\
    UnoCards (↑)                                                       & 0.20   & \cellcolor[HTML]{FFFFFF}0.81  \\
    VehiclesOpenImages (↑)                                             & 56.00  & \cellcolor[HTML]{FFFFFF}58.49 \\
    WildfireSmoke (↑)                                                  & 2.30   & \cellcolor[HTML]{FFFFFF}20.04 \\
    boggleBoards\_416x416AutoOrient\_export (↑)                        & 0.00   & \cellcolor[HTML]{FFFFFF}0.29  \\
    brackishUnderwater (↑)                                             & 3.70   & \cellcolor[HTML]{FFFFFF}1.47  \\
    dice\_mediumColor\_export (↑)                                      & 1.10   & \cellcolor[HTML]{FFFFFF}0.33  \\
    openPoetryVision (↑)                                               & 0.00   & \cellcolor[HTML]{FFFFFF}0.05  \\
    pistols\_export (↑)                                                & 49.80  & \cellcolor[HTML]{FFFFFF}66.99 \\
    plantdoc (↑)                                                       & 1.10   & \cellcolor[HTML]{FFFFFF}0.36  \\
    pothole (↑)                                                        & 17.20  & \cellcolor[HTML]{FFFFFF}25.21 \\
    selfdrivingCar\_fixedLarge\_export (↑)                             & 8.00   & \cellcolor[HTML]{FFFFFF}9.95  \\
    thermalDogsAndPeople (↑)                                           & 43.70  & \cellcolor[HTML]{FFFFFF}67.89 \\
    websiteScreenshots (↑)                                             & 0.50   & \cellcolor[HTML]{FFFFFF}1.30  \\ 
    \bottomrule
    \end{tabular}}
    \centering
    \caption{{Comparison of Grounding DINO and GLIP on ODinW. Both models are trained on O365, GoldG, and Cap4M with Swin-Tiny backbones.}}
    \label{tab:comp_odinw}
    \vspace{-.1cm}
\end{table*}

\subsection{Training Grounding DINO on RefCOCO/+/g from scratch}

To demonstrate our model's capabilities, we trained a Grounding DINO from scratch on the RefCOCO/+/g datasets, as shown in Table \ref{tab:refcoco_scratch}.
Despite time limitations that restricted us to 9 epochs of training, we achieved comparable performance to SOTA REC models. This underscores the potential of our approach to handling REC tasks effectively.

\begin{table}[htbp]
\begin{center}
    \renewcommand{\arraystretch}{1.3}
\small
\resizebox{\columnwidth}{!}{
\begin{tabular}{lllll}
\toprule
                    &       & refcoco (val/testA/testB) & refcoco+ (val/testA/testB) & refcocog (val/test) \\
\midrule
TransVG             & R101  & 81.02/82.72/78.35         & 64.82/70.70/56.94          & 68.67/67.73         \\
Grounding DINO T (9ep) & SwinT & \textbf{81.42}/85.28/76.75         & \textbf{69.14}/74.79/60.17          & \textbf{73.96}/74.37       \\
\bottomrule
\end{tabular}
}
\end{center}
\caption{Training on RefCOCO from scratch.}
\label{tab:refcoco_scratch}
\end{table}

\section{Visualizations}

\subsection{Detection Visualizations}
\begin{figure*}
    \centering
    \includegraphics[width=\linewidth]{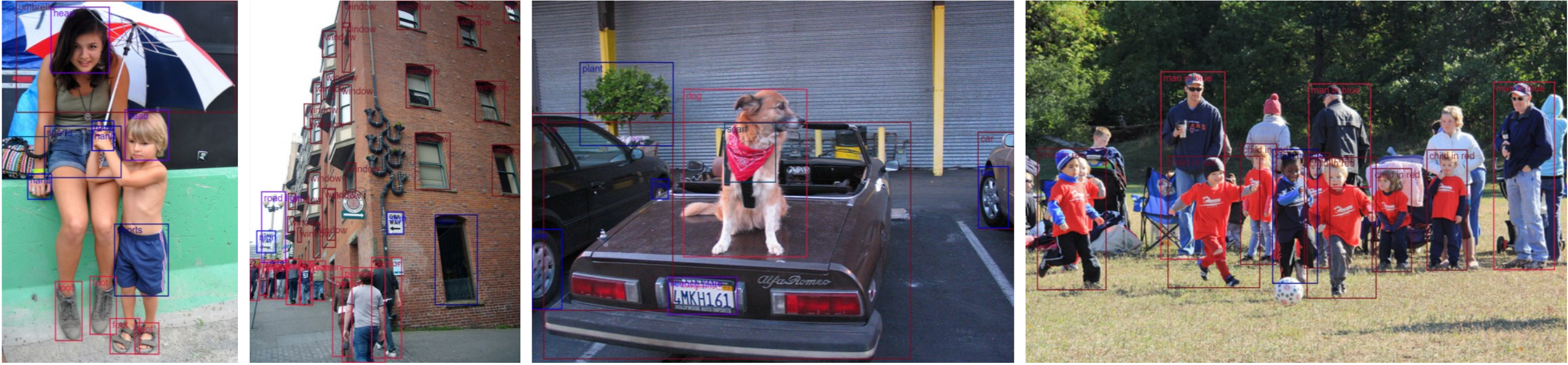}
    \caption{Visualizations of model outputs.}
    \label{fig:vis}
\end{figure*}

We present some visualizations in Fig. \ref{fig:vis}. 
Our model presents great generalization on different scenes and text inputs. 
For example, Grounding DINO accurately locates \texttt{man in blue} and \texttt{child in red} in the last image.

\subsection{Physical meaning of language-guided query }
\begin{figure}[!htbp]
\centering
 \includegraphics[width=0.9\linewidth,keepaspectratio]{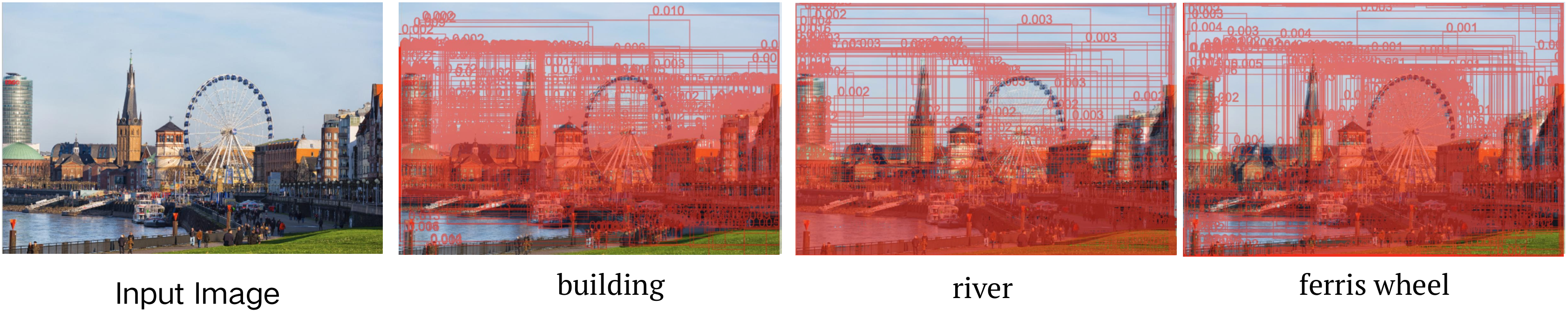}
 \caption{{Top queries in language-guided query selection.}}
 \label{fig:physical_meaning}
\end{figure}

We visualize the top 900 queries (boxes) in language-guided query selection across various prompts in Figure \ref{fig:physical_meaning}. It shows our model can have dynamic queries during inference. We will incorporate the analyses into the revised paper.

\subsection{Comparison of RefCOCO and Grounding Data}
\begin{figure}[!htbp]
\centering
 \includegraphics[width=0.9\linewidth,keepaspectratio]{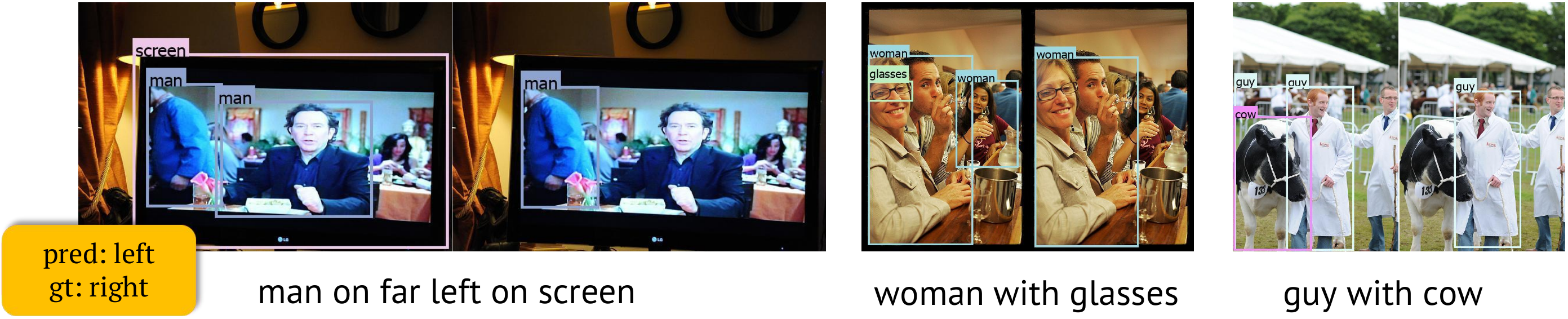}
 \caption{Our model predictions and ground-truths in RefCOCO.}
 \label{fig:refcoco_grounding}
\end{figure}

The RefCOCO has a different formulation with grounded training, resulting in a big performance gap without the RefCOCO dataset. 
As shown in Fig. \ref{fig:refcoco_grounding}, each RefCOCO text prompt corresponds to only \textit{one box}, while our model tends to predict \textit{multiple objects}.

\subsection{Marry Grounding DINO with Stable Diffusion for Object Detection and Inpainting}
We present an image editing application in Fig. \ref{fig:hero_image} (b). 
The results in Fig. \ref{fig:hero_image} (b) are generated by two processes. First, we detect objects with Grounding DINO and generate masks by masking out the detected objects or backgrounds. 
After that, we feed original images, image masks, and generation prompts to an inpainting model (typical Stable Diffusion \cite{rombach2021highresolution}) to render new images. 
We use the released checkpoints in \url{https://github.com/Stability-AI/stablediffusion} for new image generation. More results are available in Figure \ref{fig:gd_sd}.

The ``detection prompt'' is the language input for Grounding DINO, while the ``generation prompt'' is for the inpainting model.

\subsection{Marry Grounding DINO with Stable Diffusion for Object Detection and Grounded Generation}

\begin{figure*}[!htbp]
    \centering
    \includegraphics[width=\linewidth]{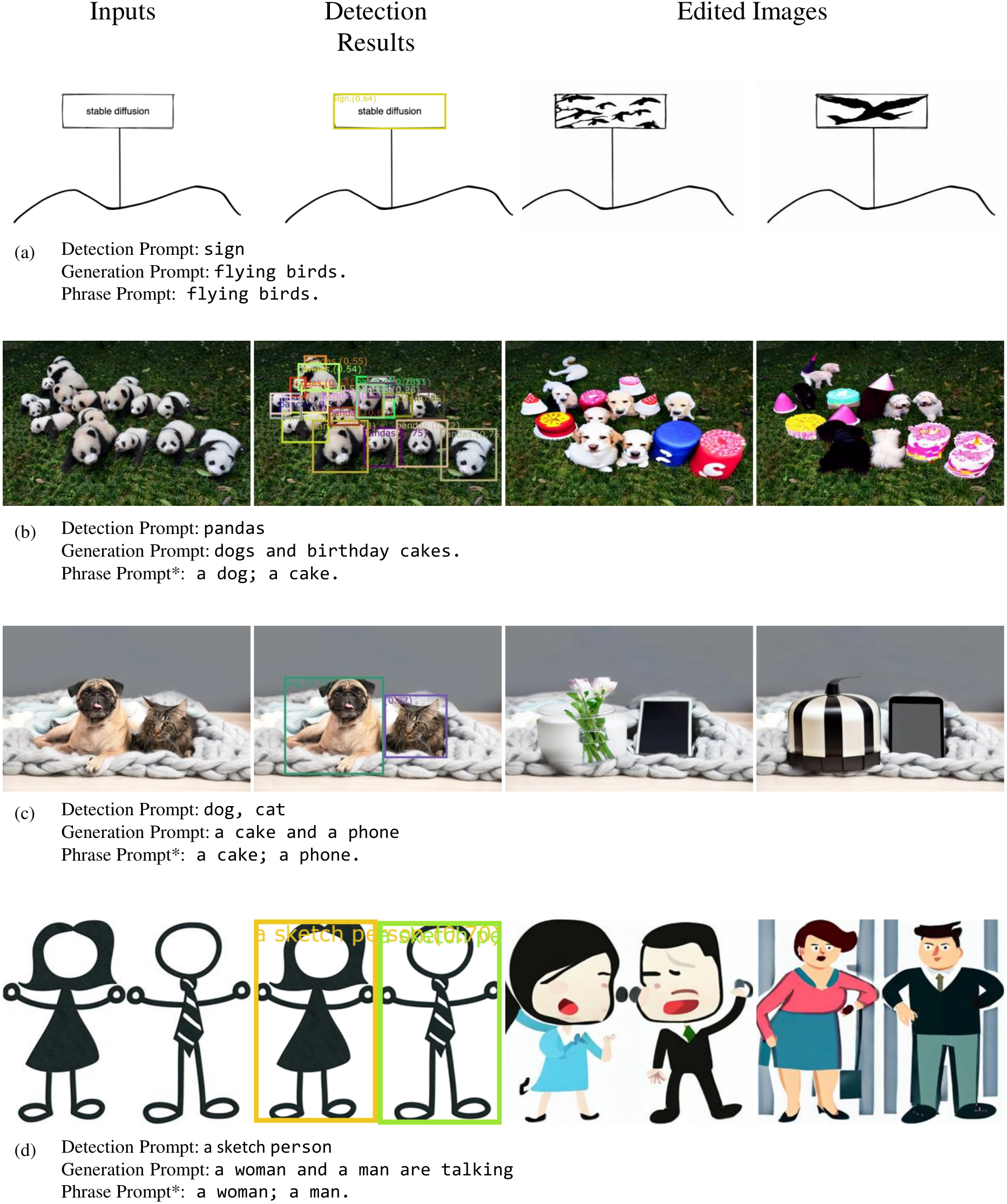}
    \caption{
    Combination of Grounding DINO and GLIGEN. We first detect objects with Grounding DINO and then perform image inpainting with GLIGEN. ``Detection Prompt'' and ``Generation Prompt'' are inputs for Grounding DINO and Stable Diffusion, respectively. ``Phrase Prompt'' are language inputs for each bounding box. The phrase prompts are separated by semicolons. *We assign phrase prompts to bounding boxes randomly. }
    \label{fig:gd_gligen}
\end{figure*}

To enable fine-grained image editing, we combine the Grounding DINO with GLIGEN \cite{GLIGEN}. 
We use the ``phrase prompt'' in Figure \ref{fig:gd_gligen} as the input phrases of each box for GLIGEN.

\begin{figure*}[!htbp]
    \centering
    \includegraphics[width=0.9\linewidth]{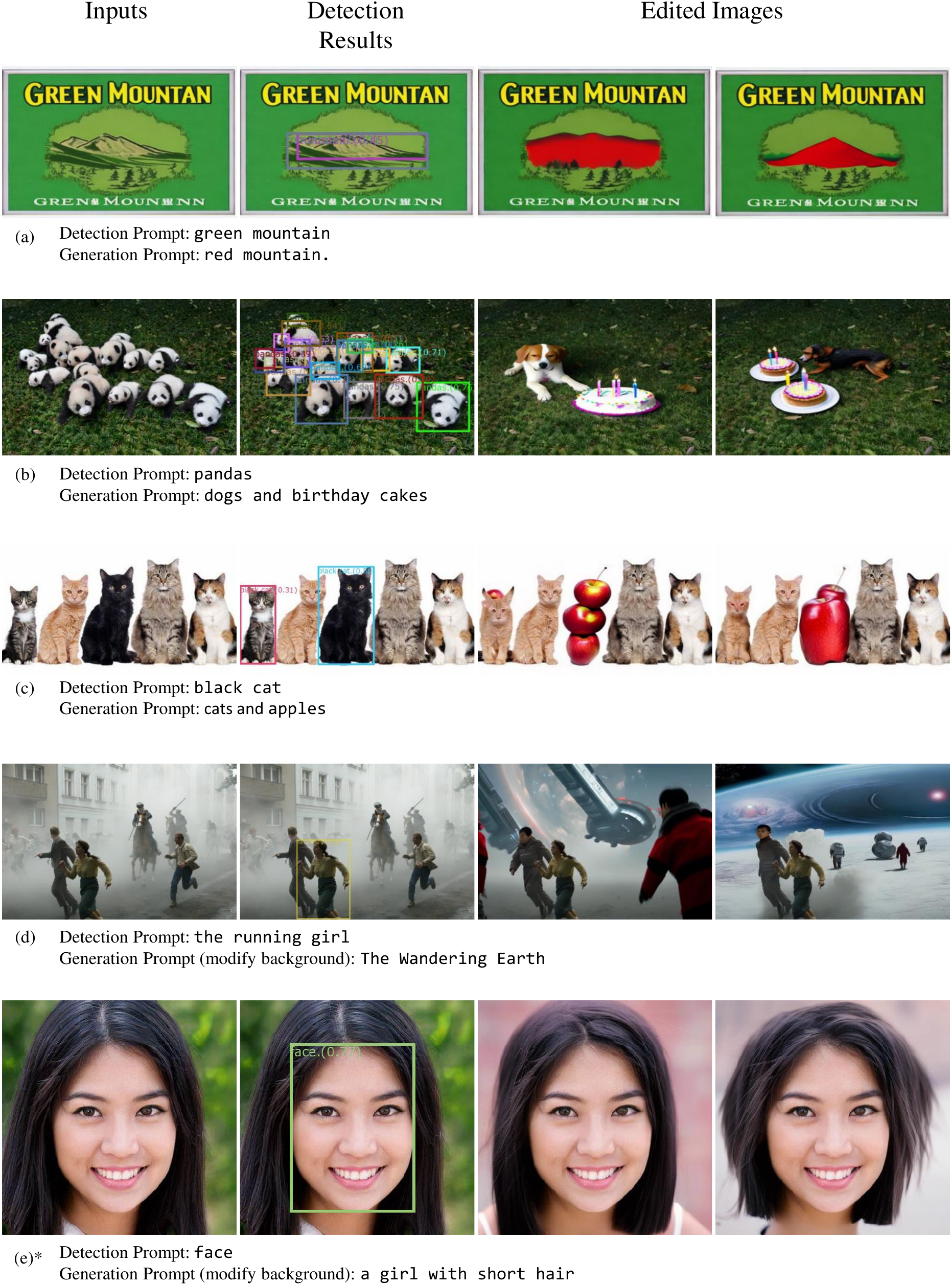}
    \caption{
    Combination of Grounding DINO and Stable Diffusion. We first detect objects with Grounding DINO and then perform image inpainting with Stable Diffusion. ``Detection Prompt'' and ``Generation Prompt'' are inputs for Grounding DINO and Stable Diffusion, respectively. *The input human face in the row (e) is generated by StyleGAN.   }
    \label{fig:gd_sd}
\end{figure*}

GLIGEN supports grounding results as inputs and can generate objects on specific positions. We can assign each bounding box an object with GLIGEN, as shown in Figure \ref{fig:gd_gligen} (c) (d). Moreover, GLIGEN can full fill each bounding box, which results in better visualization, as that in Figure \ref{fig:gd_gligen} (a) (b). For example, we use the same generative prompt in Figure \ref{fig:gd_sd} (b) and Figure \ref{fig:gd_gligen} (b). The GLIGEN results ensure each bounding box with an object and fulfills the detected regions.


\begin{table}[H]
\caption{Comparison of model size and model efficiency between GLIP and Grounding DINO.
}
\label{table:gflops}
\begin{center}
\resizebox{0.6\linewidth}{!}{
\begin{tabu}{
l@{\hskip9pt} | c@{\hskip9pt} 
c@{\hskip9pt}  c@{\hskip9pt}
}
\toprule
Model & params & GFLOPS & FPS \\
\midrule
GLIP-T \cite{li2021grounded} & 232M & 488G & 6.11  \\
Grounding DINO T (Ours) & 172M & 464G & 8.37 \\
\bottomrule
\end{tabu}
}
\end{center}
\vspace{-2mm}
\end{table}

\end{document}